\documentclass{article}
\usepackage[T1]{fontenc}
\usepackage{lineno}

\usepackage[english]{babel}
\usepackage{blindtext}

\usepackage{layout} % to visualize the page settings
\usepackage[inline]{enumitem}

\usepackage{multicol}
\usepackage{wrapfig}

\usepackage[bottom]{footmisc}

\usepackage[style=nature,backend=biber, url=false, doi=false, isbn=false]{biblatex} %numeric-comp
\usepackage{xcolor}
\definecolor{c-navy}{rgb}{0,0.08,0.45}
\definecolor{c-red}{RGB}{240,74,81}
\definecolor{c-lightblue}{RGB}{185,209,232}
\definecolor{c-blue}{RGB}{29,59,105}
\usepackage{hyperref}
\hypersetup{ %
colorlinks=true,
linkcolor=c-navy,
citecolor=c-navy,
filecolor=c-navy,
urlcolor=c-navy,
}
\usepackage{url}

\usepackage{fullpage}
\usepackage{courier}
\usepackage{authblk}
\usepackage{color}
\usepackage{graphicx}
\usepackage{colortbl}
\usepackage{tabularx}

\usepackage{array}
\usepackage{booktabs}
\usepackage{caption}
\usepackage{amsmath}
\usepackage{amsfonts}

\usepackage{tcolorbox}

\usepackage{authblk}

\usepackage{fontawesome}
\newcommand*{\emailsymbol}{{\small\faEnvelopeO}~}

\title{Adaptable Cardiovascular Disease Risk Prediction from Heterogeneous Data using Large Language Models}

\author[1,2]{Frederike L{\"u}beck \emailsymbol}
\author[2,3]{Jonas Wildberger}
\author[2,3]{Frederik Tr{\"a}uble}
\author[1,2]{Maximilian Mordig}
\author[4]{\\Sergios Gatidis$^*$}
\author[1]{Andreas Krause$^*$}
\author[1,2,3]{Bernhard Sch{\"o}lkopf$^*$}

\affil[1]{Eidgen\"ossische Technische Hochschule (ETH), Zürich, Switzerland}
\affil[2]{Max Planck Institute for Intelligent Systems, Tübingen, Germany}
\affil[3]{ELLIS Institute, Tübingen, Germany}
\affil[4]{Stanford School of Medicine, CA, USA}
\date{} % empty date

\newcommand{\jw}[1]{\textcolor{c-blue}{~Jonas: #1}}

\newcommand{\bernhard}[1]{\textcolor{red}{~B: #1}}

\begin{document}

\maketitle

\def\thefootnote{\emailsymbol}\footnotetext{Correspondence to: \href{mailto:frederike.luebeck@inf.ethz.ch}{frederike.luebeck@inf.ethz.ch}}
\def\thefootnote{*}\footnotetext{These authors contributed equally to this work and share senior authorship.
}
\def\thefootnote{\arabic{footnote}}

\vspace{3em}

%\linenumbers

\begin{abstract}

% v0.3.0

Cardiovascular disease (CVD) risk prediction models are essential for identifying high-risk individuals
and guiding preventive actions. However, existing models struggle with the challenges of real-world
clinical practice as they oversimplify patient profiles, rely on rigid input schemas, and are sensitive to distribution shifts. We developed \textsc{AdaCVD}, an adaptable CVD risk prediction framework built on large language models extensively fine-tuned on over half a million participants from the UK Biobank. In benchmark comparisons, \textsc{AdaCVD} surpasses established risk scores and standard machine learning approaches, achieving state-of-the-art performance. Crucially, for the first time, it addresses key clinical challenges across three dimensions: it flexibly incorporates comprehensive yet variable patient information; it seamlessly integrates both structured data and unstructured text; and it rapidly adapts to new patient populations using minimal additional data. In stratified analyses, it demonstrates robust performance across demographic, socioeconomic, and clinical subgroups, including underrepresented cohorts. \textsc{AdaCVD} offers a promising path toward more flexible, AI-driven clinical decision support tools suited to the realities of heterogeneous and dynamic healthcare environments.

\end{abstract}

%\begin{multicols}{2}

\section{Introduction}

Cardiovascular disease (CVD) remains the leading cause of mortality worldwide \cite{CardiovascularDiseasesCVDs}.
Accurately predicting CVD risk before symptoms manifest is an important prerequisite to initiating targeted interventions that mitigate adverse outcomes.
Current clinical guidelines \cite{rothGlobalBurdenCardiovascular2020, khanNovelPredictionEquations2023, OverviewCardiovascularDisease2023, visseren2021ESCGuidelines2021} recommend risk assessment models that estimate the future risk of CVD based on small sets of well-established risk factors.
%xThese models are typically derived from multivariate linear regression analyses on large cohort studies, often tailored to specific geographic regions.
The most prominent example is the Framingham Risk Score, a pioneering model developed from the Framingham Heart Study \cite{dagostinoGeneralCardiovascularRisk2008}, which played a key role in identifying several major risk factors.
While these base risk factors are easily measurable, less easily quantifiable factors, such as a patient’s lifestyle, also play a significant role \cite{rippeLifestyleStrategiesRisk2019}.
Several studies~\cite{alaaCardiovascularDiseaseRisk2019a, wengCanMachinelearningImprove2017} have demonstrated the potential of integrating more comprehensive health information on patients, combined with the use of stronger machine learning (ML) models that can agnostically discover the complex, non-linear interactions between a broader range of input features and disease outcomes. These methods have demonstrated increased predictive ability, enabling more personalized risk assessment.
%, in particular for subpopulations such as individuals with comorbidities~\cite{alaaCardiovascularDiseaseRisk2019}

However, both traditional medical risk scores and machine learning methods for CVD risk prediction face several critical challenges that hinder their broader clinical applicability.
First, these models rely on fixed, pre-defined, and rigid sets of input variables, restricting their ability to flexibly incorporate diverse, varying, or evolving patient information.
%When new risk factors are discovered, these models cannot easily be adapted and usually need to be completely re-estimated once large enough datasets are available. 
% require complete information
Second, they require complete and consistently formatted inputs---typically in structured, tabular forms---for accurate estimation and inference. However, in real-life clinical settings, data is often incomplete and unstructured, with documentation typically taking place in the form of clinical notes.
Third, due to their rigidity, these models are usually constrained to specific populations or environments, i.e., particular geographical regions or environments with the same data collection standards. This restriction hinders knowledge transfer between diverse healthcare environments and results in models that are poorly generalizable, for example, when facing distribution shifts.
Collectively, these limitations are in stark contrast to the demands of clinical reality. Current models lack the flexibility needed for reliable deployment in diverse healthcare settings and often exhibit inconsistent performance.
%They lead to significant variability in performance and inflexibility when using these models in practice.
%Addressing these limitations by developing CVD risk prediction models with increased adaptability could thus support broader, more effective deployment in diverse and dynamic healthcare environments.
This underscores a pressing need for more flexible, robust, and context-aware approaches to CVD risk prediction---ones that can adapt to the messy, unstructured, and heterogeneous nature of real-world clinical data.

\begin{figure*}[t]
    \centering
    \includegraphics[width=\textwidth]{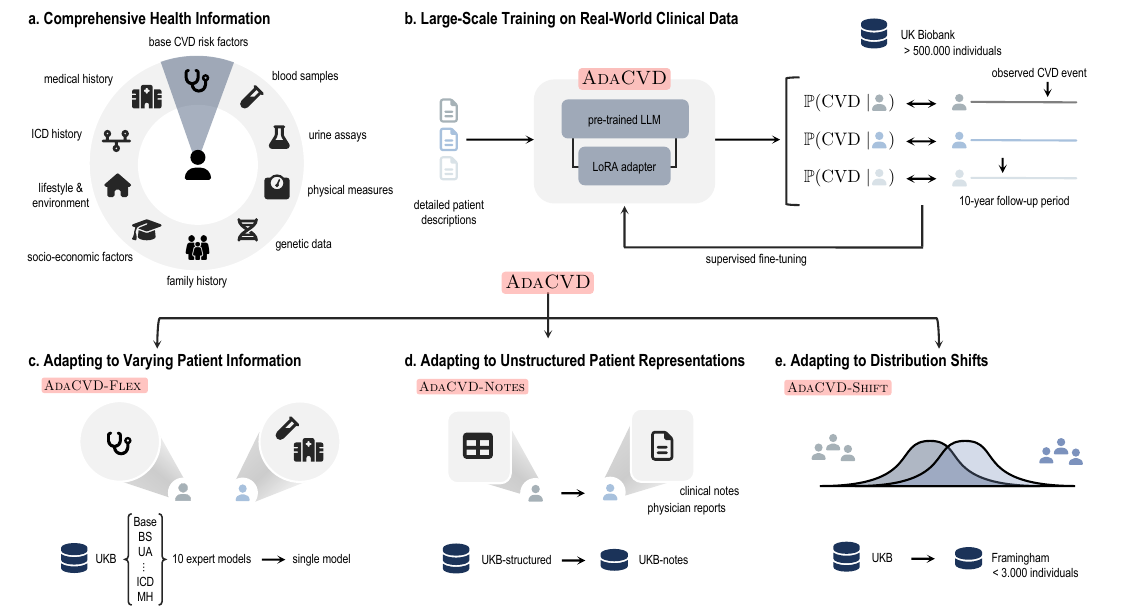}
    \caption{\textbf{Overview of our approach to adaptable cardiovascular disease (CVD) risk assessment.} \textbf{a.}~We integrate a broad spectrum of health-related information on individuals for assessing their CVD risk. Information is categorized into ten different groups.
    \textbf{b.}~\textsc{AdaCVD} is built by fine-tuning a pre-trained large language model on real-world clinical data from the UK Biobank.
    \textbf{c-e.}~We adapt \textsc{AdaCVD} along three key axes to address the challenges of real-life clinical practice.
    These are \textbf{c.}~handling varying patient information, \textbf{d.}~dealing with unstructured patient representations, and \textbf{e.}~adapting to distribution shifts.}
    \label{fig:overview}
\end{figure*}

Recently, large language models (LLMs;~\cite{bubeckSparksArtificialGeneral2023, openaiGPT4TechnicalReport2024, brownLanguageModelsAre2020}) have received significant attention for their impressive range of capabilities across multiple domains.
%These models are large-scale, general-purpose machine learning models that are trained on massive amounts of text data using self-supervised learning techniques. Pre-trained LLMs can be adapted to various downstream tasks through a multitude of model adaptation techniques, such as in-context learning or fine-tuning.
Early demonstrations of these models on clinical tasks and medical benchmarks indicate their potential to enhance or automate many aspects of clinical practice \cite{moorFoundationModelsGeneralist2023, thirunavukarasuLargeLanguageModels2023}. Notably, LLMs have been shown to encode broad clinical knowledge, allowing them to pass the US medical licensing exam with similar performance to human experts \cite{singhalLargeLanguageModels2023a, singhalExpertlevelMedicalQuestion2025}.
Further, they have demonstrated the ability to efficiently extract relevant information from clinical texts \cite{agrawalLargeLanguageModels2022} and have surpassed medical experts in clinical text summarization \cite{vanveenAdaptedLargeLanguage2024}.
The tasks analyzed thus far have primarily focused on text processing and multiple-choice question answering, yet these capabilities point to a broader and largely untapped opportunity: supporting real-world clinical decision-making by enabling models to integrate heterogeneous patient data, interpret clinical notes, and operate effectively despite missing or unstructured inputs.
%Here, we take a step toward bridging this gap by shifting the focus toward developing methods that more faithfully reflect real-world clinical workflows.

%yet their relevance to real-world clinical decision-making remains in question~\cite{boonstraArtificialIntelligenceRevolutionizing2024,wornowShakyFoundationsLarge2023, hagerEvaluationMitigationLimitations2024}. %In light of the significant limitations of existing CVD risk prediction models in handling the complexity of clinical practice, we aim to address this gap by shifting the focus toward developing methods that more faithfully reflect real-world clinical workflows.

% these tasks revolve around text processing ... We take a sligthly different perspective:

In this work, we present~\textsc{AdaCVD}, an adaptable CVD risk prediction model that extends beyond traditional risk scores by offering several important dimensions of flexibility.
We build upon the well-established pre-training and fine-tuning paradigm by using a pre-trained LLM as a starting point, and then extensively fine-tuning it on real-world data from the UK Biobank (Figs.~\ref{fig:overview}a and~\ref{fig:overview}b). The UK Biobank is one of the largest biomedical databases comprising detailed health information of approximately half a million individuals across the UK, including blood samples, medical history, lifestyle factors, and genetic information.
Our model reaches state-of-the-art performance in predicting the 10-year CVD risk of individuals, outperforming established medical risk scores and matching the performance of standard but more rigid ML models that use tabular inputs.
Importantly, what sets \textsc{AdaCVD} apart is its ability to flexibly adapt to the dynamic and often messy realities of clinical care.
We view the three critical challenges mentioned above as three distinct dimensions of clinical complexity---each reflecting a different form of distributional or contextual variation---and demonstrate that our model can address them through quick, flexible, and data-efficient adaptation.
To this end, we use our strong foundational model~\textsc{AdaCVD} as a basis to evaluate its adaptability across three distinct directions:
% serves as a foundation for ...
%adapt to overcome the following three challenges:

\begin{enumerate}[label=\roman*.]
    \item Handling varying patient information and allowing to flexibly incorporate all available information of an individual's health status, without being constrained by fixed input features (\textsc{AdaCVD-Flex}; Fig.~\ref{fig:overview}c).
    \item Dealing with diverse patient representations, from structured to unstructured patient descriptions, including clinical notes and physician reports (\textsc{AdaCVD-Notes}; Fig.~\ref{fig:overview}d).
    \item Adapting to distribution shifts when switching between heterogeneous environments (\textsc{AdaCVD-Shift}; Fig.~\ref{fig:overview}e).
\end{enumerate}

%This triad of adaptations allows~\textsc{AdaCVD} to operate effectively in diverse, dynamic, and imperfect clinical environments
%Addressing these three aspects enables learning from and adapting to diverse, heterogeneous, and dynamic healthcare environments, which supports broader and more effective deployment of data-driven CVD risk prediction models in practice.

In an extensive evaluation, we compare \textsc{AdaCVD} to relevant baselines using different levels of patient information.
We show how integrating various aspects of an individual's health status significantly improves risk assessment. In a stratified evaluation, we show that this is consistent across relevant demographic, socioeconomic, and clinical subgroups. Notably, the incorporation of detailed information is especially beneficial for elderly individuals, current smokers, individuals without formal higher education, and individuals with diabetes.
Given the importance of incorporating comprehensive patient information for risk prediction, the key challenge lies in doing so \emph{flexibly}, while accommodating data variability and incompleteness.
We demonstrate that \textsc{AdaCVD} can handle heterogeneous and incomplete patient inputs at inference time, effectively leveraging all available information without being constrained by a fixed input schema.
Beyond flexibility with respect to the input content, we further demonstrate flexibility with respect to input format.
Although the model is extensively fine-tuned on structured patient representations, we show that it can transfer to unstructured textual formats, such as clinical notes and physician reports. Notably, this transfer is highly data-efficient, requiring up to 100 times fewer examples compared to training from scratch.
Lastly, when evaluated on a patient cohort from the Framingham Heart Study \cite{FraminghamHeartStudy}, \textsc{AdaCVD} demonstrates robustness and strong adaptability to distribution shifts.

% We demonstrate that our model can flexibly accommodate heterogeneous and incomplete patient information at inference time, enabling it to effectively incorporate all available data—whether structured, unstructured, or partially missing—without requiring a fixed input schema. This adaptability allows \textsc{AdaCVD} to operate robustly across a range of real-world clinical scenarios, where the availability and completeness of patient data can vary widely. Massively fine-tuned on structured patient representations, our model is able to transfer to unstructured, textual representations of patients, such as those found in clinical notes and physician reports. This transfer process is data-efficient, requiring up to 100 times fewer data points from the target domain than when learning from scratch.

%\clearpage
%\input{content/02_method}
%\clearpage
\section{Results}

\subsection*{Constructing Patient Representations and Model Training}

\begin{figure*}[t!]
    \centering
    \includegraphics[width=\textwidth]{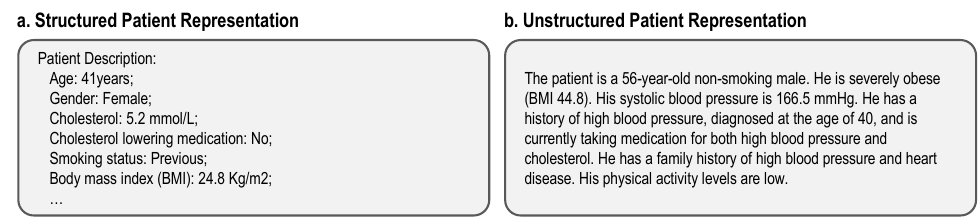}
    \caption{
    \textbf{Overview of the patient representations used for training and evaluating \textsc{AdaCVD}.} We generated two types of patient descriptions of over half a million individuals using real-world data from the UK Biobank and the Framingham Heart Study.
    \textbf{a.}~Example of a structured patient description, where patient information is serialized into a predefined text format.
    \textbf{b.}~Example of a free-text patient summary, simulating clinical notes. These representations are not standardized, capturing the variability found in real-world clinical documentation.
    }
    \label{fig:dataset}
\end{figure*}

%The defining strength of LLMs is their ability to process and reason over natural language text, which is especially valuable in clinical settings where text is the dominant form of documentation and information exchange.
%A key goal of this work is to model a realistic clinical scenario, which requires constructing a representative evaluation framework and generating patient-level data that mirrors the variability found in practice.
In the clinic, natural language text is the dominant form of documentation and information exchange.
Clinical notes, physician reports, and discharge summaries typically contain detailed textual descriptions of patients, capturing a wide range of observations, diagnostics, and contextual information.
These patient descriptions are rich in information but pose challenges to risk prediction models as they move away from clearly defined and structured sets of input features. Yet this variability reflects the demands of real-world clinical practice: Given \emph{any} information on a patient---regardless of structure or format---we are interested in understanding their risk of developing cardiovascular disease.

To train \textsc{AdaCVD} on realistic clinical inputs, we generated patient descriptions of over half a million individuals using data from the UK Biobank \cite{UKBiobankUK2025} and the Framingham Heart Study \cite{FraminghamHeartStudy}. We used two different approaches, resulting in two types of patient representations of increasing complexity.
In the first version, we serialized patient data into a highly structured text format (see Fig.~\ref{fig:dataset}a). In the second version, we generated free-text patient summaries (see Fig.~\ref{fig:dataset}b) that simulate clinical notes (see \hyperref[supp:method]{Methods} for details).
The patient information used to create these descriptions varies across experiments throughout this work. Established CVD risk prediction models typically rely on a small set of well-known risk factors, such as age, cholesterol levels, blood pressure, smoking status, and diabetes; referred to here as the base risk factors. Beyond these, we defined nine additional information categories that capture broader aspects of patient health: Lifestyle \& Environment, Sociodemographic factors, Physical Measures, Urine Assays, Blood Samples, Family History, Polygenic Risk Scores, ICD Codes, and Medical History.
%Figure \ref{fig:dataset}a shows the lengths of patient descriptions as measured in the number of tokens\footnote{A token refers to a unit of text used by language models during processing, which typically corresponds to a word or subword (e.g., common word fragments, punctuation marks, or individual characters in rare cases); see \hyperref[supp:method]{Methods} for further details.} across the two underlying data sources (UK Biobank and Framingham), the two prompt formats (structured and free-text), and varying amounts of input information.

We developed our model \textsc{AdaCVD} to flexibly reason over these diverse patient representations.
Our approach follows the well-established pre-training and fine-tuning paradigm, utilizing a pre-trained transformer-based LLM with general language understanding capabilities as a foundation and tailoring it to the specific task of CVD risk prediction. Specifically, we fine-tuned \texttt{Mistral-7B-Instruct} \cite{jiangMistral7B2023} on over $467\,063$ structured patient representations from the UK Biobank (\texttt{UKB-structured}) to predict the 10-year risk of developing CVD. 
To reduce computational demands, we used low-rank adaptation (LoRA; \cite{huLoRALowRankAdaptation2021}), a parameter-efficient fine-tuning method that updates only a small fraction of model weights.

%We frame the CVD risk prediction task as a binary classification problem in the token space (similar to \cite{hegselmannTabllmFewshotClassification2023, belyaevaMultimodalLLMsHealth2023a}). Instead of producing a numerical risk prediction in text form, we extract the likelihood of the model answering \textit{Yes} or \textit{No} to a question posed in binary form: \emph{Will this patient experience a major cardiovascular event within the next ten years?} These probabilities are subsequently normalized to generate the final CVD risk prediction. For training, we completed the prompt with a binary label based on the true observed 10-year CVD outcome of each participant and learned the parameters to minimize the cross-entropy loss between predicted probabilities and observed outcomes. To reduce computational costs, we used Low-Rank Adaptation (LoRA), a parameter-efficient fine-tuning technique with which we updated only about 0.13\% of the model's parameters. Further details on the training process are provided in Methods.

\subsection*{Benchmarking Performance Using Base Risk Factors}

% AdaCVD reaches state-of-the-art performance
% AdaCVD accurately predicts CVD risk

\begin{figure*}[t!]
    \centering
    \includegraphics[width=\textwidth]{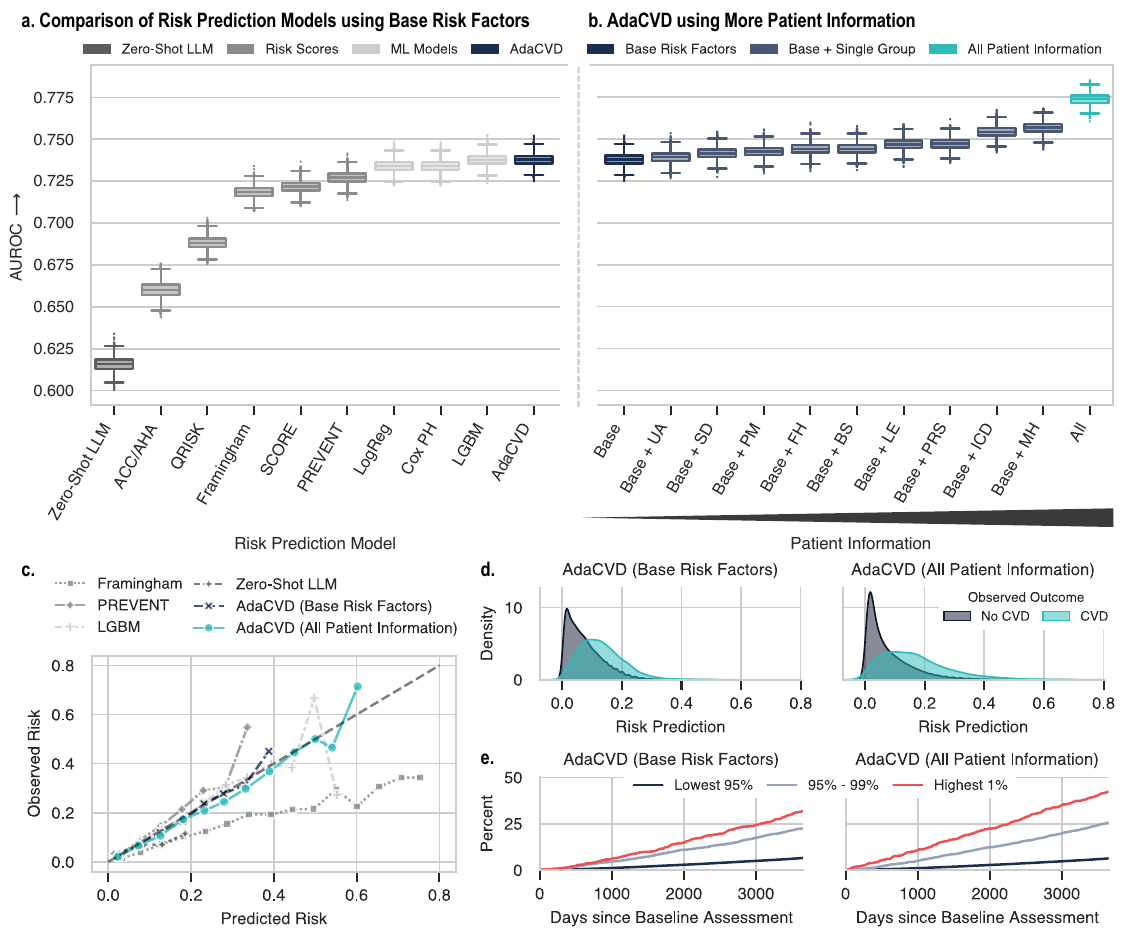}
    \caption{\textbf{Evaluation of \textsc{AdaCVD}.}
    \textbf{a.}~Comparison of CVD risk prediction models using only a limited set of base risk factors. Predictive performance is measured by the area under the receiver operating characteristic curve (AUROC; y-axis), reflecting the models’ ability to distinguish between individuals who develop CVD and those who do not. \textsc{AdaCVD} outperforms established medical risk scores and machine learning baselines. Zero-shot LLMs perform poorly.
    \textbf{b.}~Predictive performance of \textsc{AdaCVD} improves as additional patient information is incorporated.
    Acronyms: Urine Assays (UA), Sociodemographic factors (SD), Physical Measures (PM), Family History (FH), Blood Samples (BS), Lifestyle \& Environment (LE), Polygenic Risk Scores (PRS), ICD Codes (ICD), and Medical History (MH). Each feature group includes the base risk factors. The \textit{All Patient Information} setting integrates all feature categories.
    \textbf{c.}~Calibration plot comparing predicted and observed risk across binned risk strata. The diagonal line indicates perfect calibration. \textsc{AdaCVD} model's using different inputs are all well-calibrated.
    \textbf{d.}~Risk prediction distributions for individuals that developed CVD vs. those who did not, using either base risk factors (left) or all patient information (right). Distributions become more separable with comprehensive inputs.
    \textbf{e.}~Event curves stratified by predicted risk percentiles (lowest $95\%$, $95-99\%$, and highest $1\%$) when using only the base risk factors (left) and when incorporating all patient information (right). The x-axis denotes the 10-year follow-up period (in days); the y-axis shows the observed incidence for each risk group up until that day; the rightmost points indicate the observed 10-year incidence rates for each risk group. The event curves show clearer separation and improved risk stratification when using all patient information.
    }
    \label{fig:panel-auroc}
\end{figure*}

We first evaluated model performance in a controlled setting using only the base risk factors that are commonly used by established CVD risk scores.
These base risk factors include age, gender, smoking status, diabetes, total and HDL cholesterol, cholesterol medication use, blood pressure, blood pressure medication use, body mass index (BMI), ethnic background, and estimated glomerular filtration rate (eGFR).
%These features are well-known risk factors for CVD and are routinely collected in clinical practice. To enable direct comparison to baseline models, we trained our model using the structured version of the patient representations (UKB-tab) that contained only these base risk factors.

\textsc{AdaCVD} achieved state-of-the-art performance for CVD risk prediction when using only the base risk factors, with an area under the receiver operating characteristic curve (AUROC) of 0.738 (Fig.~\ref{fig:panel-auroc}a). The predictions are well-calibrated (Fig.~\ref{fig:panel-auroc}c), meaning that the predicted risk matched the observed risk.
Its performance matched the performance of gradient-boosted trees (AUROC: 0.738), which are specifically designed for and known to excel with this setting of structured tabular inputs.
Performance was superior to simpler ML models, including logistic regression and the Cox proportional hazards model.
Notably, all ML models surpassed established medical risk scores, which showed great variability in performance (Fig.~\ref{fig:panel-auroc}a). Among these scores, PREVENT \cite{khanDevelopmentValidationAmerican2024} achieved the highest AUROC (0.727); QRISK \cite{hippisley-coxDerivationValidationQRISK2007} and ACC/AHA \cite{arnett2019ACCAHA2019} were significantly worse (AUROC 0.688 and 0.660).

Across five open-access LLMs of small (2-3B parameters) and medium size (7-8B parameters), post-fine-tuning performance was similar for models of comparable sizes, and predictions were highly correlated (Supplementary Fig.~\ref{fig:llms-eval}a–b). In contrast, zero-shot LLMs performed poorly (AUROC: 0.616) and inconsistently followed instruction formats (Fig.~\ref{fig:panel-auroc}a and Supplementary Fig.~\ref{fig:llms-eval}a), emphasizing the relevance of domain-specific fine-tuning.

\subsection*{Incorporating Broader Patient Information Improves Risk Prediction}

A major limitation of established CVD risk scores is that they incorporate only a limited set of input features for risk prediction. While these base risk factors are essential, they do not capture the full complexity of an individual’s health profile. In particular, it is well known that other factors, such as a patient's lifestyle, also play a significant role \cite{rippeLifestyleStrategiesRisk2019}.

To assess the relevance of broader patient information, we evaluated models trained with additional feature groups beyond the base risk factors.
Figure~\ref{fig:panel-auroc}b shows the performance improvements with the inclusion of additional patient information.
Compared to using only the base risk factors (AUROC 0.738), incorporating all patient information improves risk prediction by 4.9\% (AUROC 0.774). Feature groups with the largest individual performance gains included lifestyle \& environment (+1.23\%), polygenic risk scores (+1.29\%), ICD codes (+2.25\%), and medical history (+2.58\%).

Using all patient information led to a broader and more distinct separation in the distribution of predicted risks between individuals who did and did not develop CVD (Fig.~\ref{fig:panel-auroc}d).
Furthermore, risk stratification based on predicted percentiles showed enhanced separation of event curves (Fig.~\ref{fig:panel-auroc}e), which implies improved identification of high-risk individuals.

The patient description lengths varied across feature groups and individuals. % (Fig.~\ref{fig:dataset}a).
An interesting comparison arises between two groups: ICD codes and medical history. Although both represent similar clinical content, they differ significantly in structure and length: ICD codes are concise lists of standardized codes (median length: 201 tokens\footnote{A token refers to a unit of text used by language models during processing, which typically corresponds to a word or subword (e.g., common word fragments, punctuation marks, or individual characters in rare cases).}), while medical history consists of longer, self-reported, questionnaire-style narratives (median length: 469 tokens). Despite these differences, both formats demonstrated comparably strong predictive performance, indicating that \textsc{AdaCVD} can effectively extract clinical signals from both highly standardized, sparsely occurring codes and more naturalistic, narrative inputs.

%\bernhard{, indicating that LLMs may help alleviate the need to ask medical personnel to enter data in structured form.}

\subsection*{Enhanced Risk Assessment Across Demographic, Clinical, and Socioeconomic Subgroups}
%\subsection*{Enhanced Risk Assessment for Underrepresented Subgroups}

\begin{figure*}[t!]
    \centering
    \includegraphics[width=\textwidth]{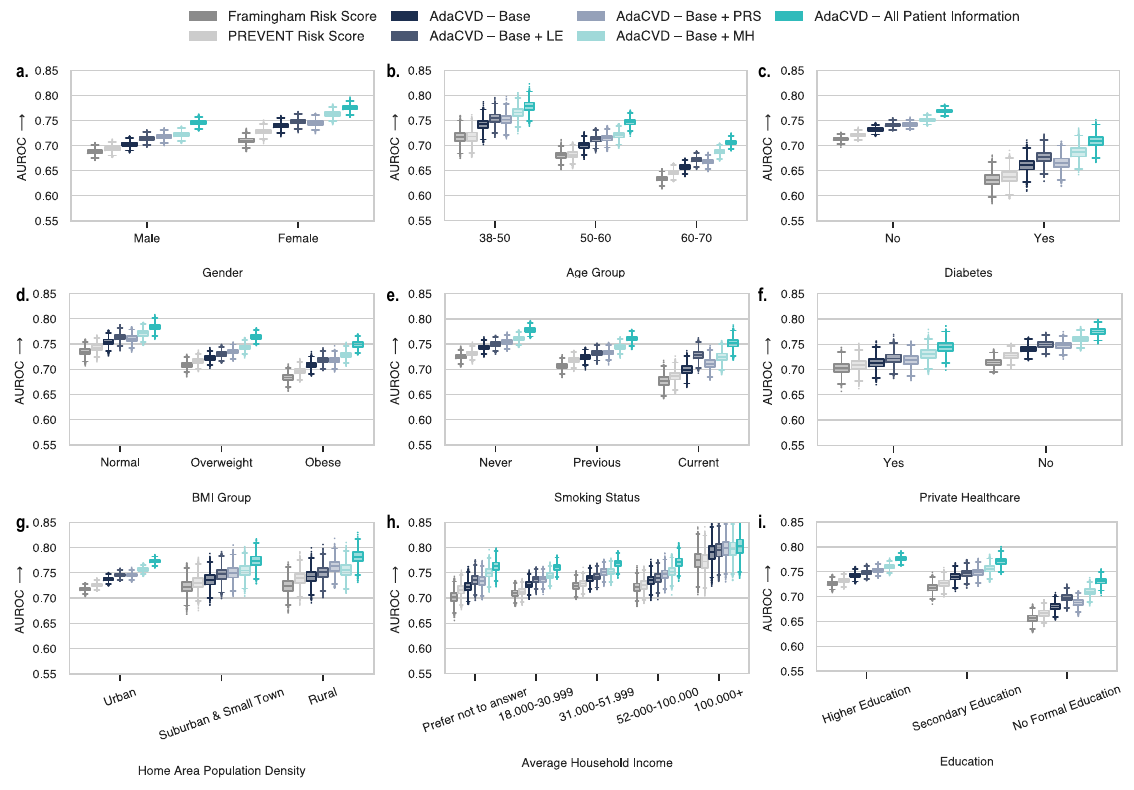}
    \caption{
    \textbf{Stratified evaluation across subgroups.}
    Evaluation of model performance (AUROC; y-axis) across \textbf{(a, b)} demographic, \textbf{(c-e)} clinical, and \textbf{(f-i)} socioeconomic subgroups. Each panel compares the Framingham Risk Score, PREVENT Risk Score, and five variants of \textsc{AdaCVD}, from using only base risk factors to incorporating all available patient information. Across all subgroups, performance improves with the inclusion of additional patient information. The largest gains are observed for elderly individuals (+7.55\%; \textbf{b}), those without higher education (+7.50\%; \textbf{i}), current smokers (+7.45\%; \textbf{e}), and individuals with diabetes (+7.21\%; \textbf{c}). Despite subgroup variation, the relative ranking of feature groups remains largely consistent.
    }
    \label{fig:stratified}
\end{figure*}

To assess potential biases of LLM-based models, we evaluated model performance across demographic, socioeconomic, and clinical subgroups (Fig.~\ref{fig:stratified}).
All subgroups benefited from the inclusion of broader patient information. Notably, certain groups experienced even more pronounced performance gains.
We measure the relative performance gain in percent between the model using only the base risk factors and the model using all patient information. 
The largest performance improvements were observed in elderly individuals (+7.55\%), individuals without higher education (+7.50\%), current smokers (+7.46\%), and individuals with diabetes (+7.21\%).
Across all subgroups, the ranking of feature groups remained consistent, with only minor exceptions, indicating that including comprehensive health information does not come at the expense of certain subpopulations.

\subsection*{Adapting to Incomplete and Variable Patient Information}

\begin{figure*}[t!]
    \centering
    \includegraphics[width=0.95\textwidth]{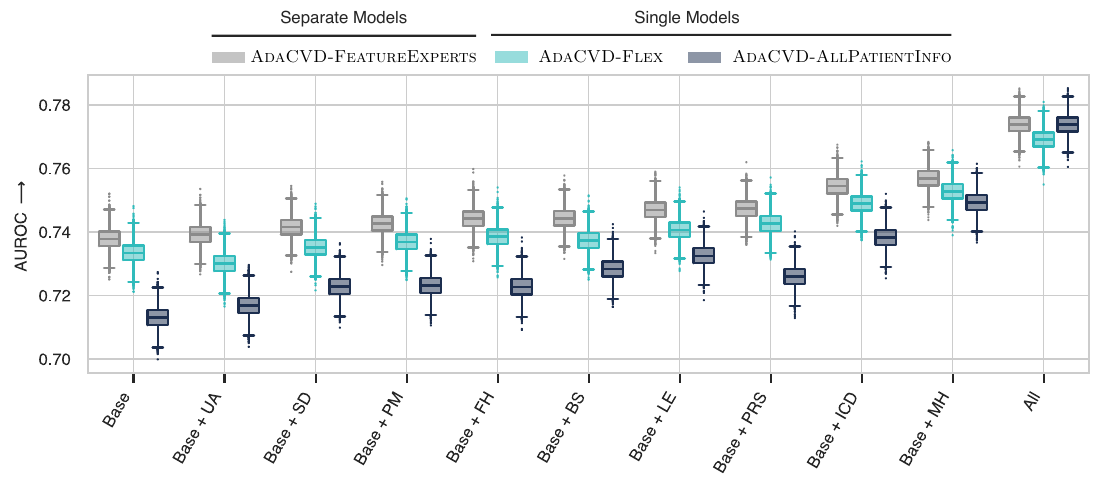}
    \caption{\textbf{Evaluation of \textsc{AdaCVD} and \textsc{AdaCVD-flex} when dealing with varying patient information.}
    Evaluation of model performance (y-axis; AUROC) when different amounts of patient information are available at inference time (x-axis; feature groups).
    The grey boxes (\textsc{AdaCVD-FeatureExperts}) represent the performance of feature-specific expert models trained separately for each feature group, serving as an upper bound.
    The dark blue boxes (\textsc{AdaCVD-AllPatientInfo}) show the performance of the model trained on all features (single model) when evaluated on partial inputs.
    The turquoise boxes show the performance of \textsc{AdaCVD-Flex}, a version of the model further fine-tuned on patient representations with varying subsets of information.
    While slightly below the expert upper bound, \textsc{AdaCVD-Flex} provides robust and accurate predictions across all input conditions, enabling flexible deployment in real-world settings with variable and incomplete data.
    }
    \label{fig:sep-flex}
\end{figure*}

In the above experiments, we operated under ideal conditions where all patient records contained the same set of complete data features.
While this is a common assumption in machine learning research, real-world clinical data rarely meets this standard. Patient records are often incomplete or inconsistent, with varying levels of information available for different patients.
Therefore, an important requirement for deploying CVD risk prediction models in practice is their ability to handle \emph{varying} patient data at inference time---that is, to make predictions based on whatever information is available for a given patient.
A key advantage of using LLM-based models is their flexibility in handling such variability: Patient information can be represented as free-text descriptions, allowing the model to process diverse and non-standardized inputs without relying on fixed feature sets and imputation of missing values.
To evaluate whether a single model can flexibly handle varying levels of patient information at inference time, we compared three approaches. As a performance upper bound, we used the 11 expert models, each trained on a specific feature group. 
Next, we evaluated the model trained on all available features (\textsc{AdaCVD-AllPatientInfo}), testing its robustness when given incomplete information at inference. While this model maintained reasonable performance under missing input conditions, its accuracy was noticeably reduced (Fig.~\ref{fig:sep-flex}).

To improve robustness, we leveraged the model’s adaptability by further fine-tuning it on patient descriptions with varying levels of detail.
In each training instance, only a random subset of features was included. This adaptation yielded a flexible version of the model, \textsc{AdaCVD-Flex}, capable of making accurate predictions from any available information. While its performance remains slightly below that of the specialized expert models trained on fixed feature subsets, the difference is small (Fig.~\ref{fig:sep-flex}).
Crucially, the slight reduction in predictive accuracy is outweighed by the significant clinical advantage of having a \emph{single} model that can seamlessly handle varying amounts of patient information.

\subsection*{Adapting to Textual Patient Representations}

Textual representations of patients---such as those found in clinical notes, discharge reports, or physician summaries---are among the most common data modalities in clinical settings but cannot be processed directly by conventional risk scores or machine learning models. Such texts are diverse and significantly less structured than the prompts we used above (\texttt{UKB-structured}) to train \textsc{AdaCVD}.
We evaluated our model's ability to adapt to these unstructured representations using the generated free-text patient descriptions \texttt{UKB-notes}. While these LLM-generated summaries mostly preserved key clinical information, they often expressed it in a more abstract or inferred form. For example, exact BMI values were sometimes replaced with phrases such as \textit{``the patient is obese"}, and detailed physical activity metrics were summarized as \textit{``the patient is very active."} The focus of this work does not lie in evaluating these summaries.
Instead, we treat them as given and assess how efficiently our model trained on structured representations could adapt to this unstructured input format. 

\begin{figure*}[t!]
    \centering
    \includegraphics[width=\textwidth]{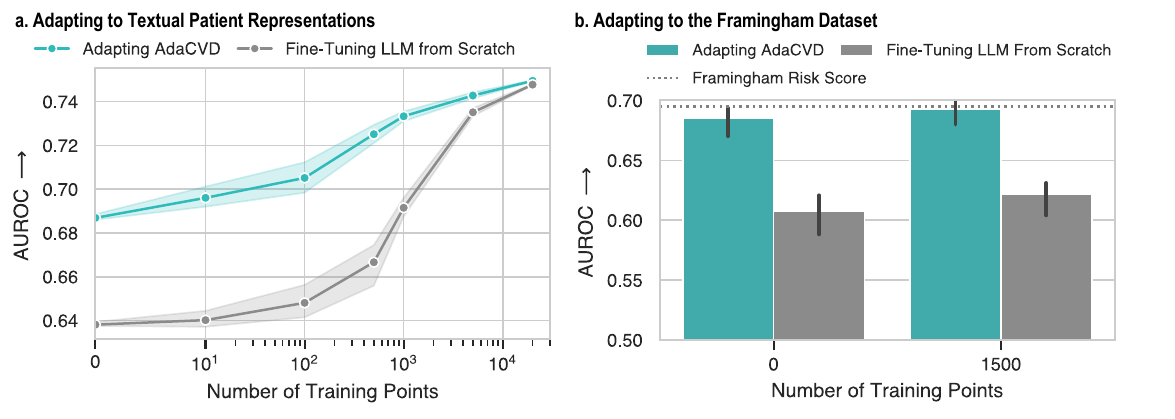}
    \caption{
    \textbf{Data-efficient adaptation of \textsc{AdaCVD} to new input formats and populations.} 
    \textbf{a.}~Adaptation to unstructured, textual patient representations. The turquoise curve shows the performance of \textsc{AdaCVD-Text}, adapted from a model trained on structured inputs, while the grey curve represents a model fine-tuned from scratch. \textsc{AdaCVD-Text} consistently outperforms the model trained from scratch, especially in low-data regimes.
    \textbf{b.}~Adaptation to a shifted target population (Framingham cohort). In both the zero-shot and few-shot settings (1500 training examples), the adapted \textsc{AdaCVD} model (turquoise bars) outperforms the baseline LLM fine-tuned from scratch (grey bars), and approaches the performance of the original Framingham Risk Score (dotted line).
    }
    \label{fig:adaptation}
\end{figure*}

We compared two strategies: (i) fine-tuning an LLM from scratch on the free-text representations and (ii) continuing fine-tuning our structured-data model \textsc{AdaCVD}, resulting in an adapted version \textsc{AdaCVD-Text}.
Figure~\ref{fig:adaptation}a shows the AUROC as a function of the number of training examples. 
The adapted model \textsc{AdaCVD-Text} consistently outperformed the model fine-tuned from scratch, particularly in low-data regimes. Even without any data from the target domain, it performed well (AUROC 0.685), and with just 10 examples, it reached an AUROC of 0.697.
In contrast, fine-tuning an LLM from scratch required 100 times more data ($>1000$ points) to achieve similar results.

\subsection*{Adapting to Distribution Shifts}

Beyond changes in patient representation, another critical challenge for deploying risk prediction models in practice is adapting to distribution shifts. Such shifts can arise, for example, when applying models across different hospitals or geographic regions.
%These shifts can lead to distributional changes in both risk factors and disease outcomes. 

To assess the adaptability of LLMs in this scenario, we used a publicly available subset from the original Framingham Heart Study \cite{FraminghamHeartStudy, TeachingDatasetsPublic}, conducted in the 1960s in Framingham, USA.
Figure \ref{fig:ukb-fram} in the Supplementary Material compares the distribution of the base risk factors and disease outcomes between the two cohorts. The marginal distributions of these two datasets differ, likely for two main reasons. First, geographical differences between the US and the UK contribute to variations in risk factors. Second, temporal differences play a significant role: the Framingham study was the first landmark investigation that identified major CVD risk factors, which influenced both patient outcomes and preventive care strategies over time.

We evaluate \textsc{AdaCVD} in a zero-shot setting and after further fine-tuning it on a small dataset from the target domain ($n=1500$). For comparison, we also assess the base LLM, both zero-shot and after fine-tuning on the same limited dataset. Figure \ref{fig:adaptation}b shows the AUROC computed on a validation set of the Framingham dataset.
When applied zero-shot, \textsc{AdaCVD} already achieved strong performance (AUROC 0.687). Similar to the findings above, the model is able to adapt quickly and data-efficiently to this new patient cohort, ultimately reaching the performance of the original Framingham risk score (AUROC 0.693), which was derived from a larger version of this dataset.
In contrast, the base LLM performed poorly in both the zero-shot (AUROC 0.614) and fine-tuned setting (AUROC 0.625).
%\clearpage
\section{Discussion}

% To add:
% Real-world scenarios combine the three axes that we have considered separately (disentangled)

Cardiovascular disease risk estimation is a critical task in preventive healthcare, yet the tools that clinicians rely on often fall short when confronted with the complexities of real-world clinical settings. To address these limitations, we developed \textsc{AdaCVD}, an adaptable risk prediction model based on a large language model (LLM) fine-tuned on diverse patient data from the UK Biobank.
\textsc{AdaCVD} moves beyond the rigidity of existing approaches and is designed to be robust to the diverse, dynamic, and often imperfect nature of clinical data.

Key findings and their implications should be emphasized.
First, in a benchmark setting with structured, complete, and consistent data, \textsc{AdaCVD} achieved state-of-the-art performance in 10-year CVD risk prediction. It outperformed established medical risk scores and matched the performance of specialized tabular machine learning models that are specifically optimized for this setting. 
However, such idealized data conditions rarely occur in practice. Clinical data is often incomplete, heterogeneous, and inconsistently formatted.
Especially when considering comprehensive patient information---which we have shown to be highly relevant---consistency is difficult to ensure.
%including detailed medical histories, comorbidities, and lifestyle factors.
In these more realistic scenarios, we showed that \textsc{AdaCVD} maintained robust predictive performance, due to its ability to handle free-text inputs and variable formats. Existing approaches typically fail under these conditions for their reliance on fixed input formats and assumptions of data completeness.
Our stratified analyses further demonstrated that this flexibility of integrating comprehensive patient information benefits underrepresented groups, including elderly individuals, smokers, individuals with diabetes, and individuals without higher education.
Across all groups, we observed clear benefits from the inclusion of detailed health-related information, showing that this does not come at the expense of specific subpopulations.

Second, this work also highlights the untapped potential of using clinical notes for ML-based decision support. Free-text documentation is ubiquitous in the clinic and captures rich, context-specific insights about a patient’s condition that are often lost in rigid tabular representations. We demonstrated that \textsc{AdaCVD} can reason effectively over such unstructured input and that fine-tuning the model on structured data facilitates this capability. This finding opens up exciting possibilities for applying similar methods to other conditions beyond cardiovascular disease, potentially enabling opportunistic screening and decision support without the need for costly, time-consuming data preprocessing or manual feature engineering.
%It paves the way for more scalable, flexible systems that better integrate into the natural workflows of clinical practice.

Third, a key strength of \textsc{AdaCVD} is its adaptability across three critical dimensions: input content, input format (structured data vs. free-text), and population distribution.  Our results show that the adaptation of \textsc{AdaCVD} along these axes is data-efficient, requiring up to 100 times fewer labeled examples to adapt to new settings compared to models trained from scratch. This capability is highly relevant for real-world applicability, where large, labeled datasets from the target setting may be unavailable or costly to obtain.
We have explicitly disentangled these three axes to understand their individual effects on model performance. While real-world clinical deployments will likely involve simultaneous shifts across multiple dimensions, we believe our findings generalize well to these combined scenarios.

Finally, our study emphasizes the importance of task-specific fine-tuning for achieving high performance on complex clinical decision-making tasks. While general-purpose LLMs have demonstrated impressive capabilities in medical question answering \cite{singhalLargeLanguageModels2023, singhalExpertlevelMedicalQuestion2025} and summarization \cite{vanveenAdaptedLargeLanguage2024}, we observed that open-access models of small and medium size performed poorly in CVD risk prediction when applied zero-shot. This underscores the need for targeted fine-tuning using task-specific data. Although prior work \cite{hanEvaluationGPT410year2024a} has shown that ChatGPT-4, a significantly larger model, performs comparably to the Framingham risk score for CVD risk prediction in zero-shot settings, our domain-specific fine-tuning approach achieved higher accuracy using significantly smaller models (with approximately 300 times fewer parameters), which can be run on modest hardware. Moreover, our model can be deployed locally without relying on external APIs, and thereby ensures that patient data remains private and compliant with regulatory standards.

This study has limitations. First, model training was conducted exclusively on data from the UK Biobank, which may not fully represent the diversity of the global population. Although we included evaluation on the Framingham cohort, further validation across different countries and healthcare systems is necessary.
Second, there is a lack of real-world datasets that pair textual patient representations, such as clinical notes and physician reports, with subsequent disease outcomes. To address this gap, we synthetically generated free-text patient descriptions to simulate clinical notes using structured real-world inputs. While previous work has demonstrated that such synthetic summaries can be both accurate and clinically relevant~\cite{agrawalLargeLanguageModels2022, vanveenAdaptedLargeLanguage2024}, we emphasize the need for the creation of publicly available, high-quality datasets containing real clinical text. Ideally, these notes should be collected longitudinally to support research on disease progression and long-term outcomes.

%Building on our findings, future work could explore extending our approach to integrating multi-modal inputs such as medical images, e.g., using Vision-Language Models.

In summary, our work presents a compelling new pathway for CVD risk prediction, showing that large language models, when fine-tuned on population-scale clinical data, can effectively support clinical decision-making and address the complexities of real-world healthcare. By enabling flexible integration of diverse patient information and robust adaptation to new settings, this framework offers a promising route toward increased interoperability and knowledge transfer across heterogeneous clinical environments.

\clearpage
%\newpage
\printbibliography

@inproceedings{agrawalLargeLanguageModels2022,
 author = {Monica Agrawal and Stefan Hegselmann and Hunter Lang and Yoon Kim and David Sontag},
 booktitle = {Proceedings of the 2022 Conference on Empirical Methods in Natural Language Processing},
 conference = {EMNLP},
 title = {Large Language Models are Few-Shot Clinical Information Extractors},
 url_paper = {https://arxiv.org/pdf/2205.12689.pdf},
 year = {2022}
}

@inproceedings{brownLanguageModelsAre2020,
 author = {Brown, Tom and Mann, Benjamin and Ryder, Nick and Subbiah, Melanie and Kaplan, Jared D and Dhariwal, Prafulla and Neelakantan, Arvind and Shyam, Pranav and Sastry, Girish and Askell, Amanda and Agarwal, Sandhini and Herbert-Voss, Ariel and Krueger, Gretchen and Henighan, Tom and Child, Rewon and Ramesh, Aditya and Ziegler, Daniel and Wu, Jeffrey and Winter, Clemens and Hesse, Chris and Chen, Mark and Sigler, Eric and Litwin, Mateusz and Gray, Scott and Chess, Benjamin and Clark, Jack and Berner, Christopher and McCandlish, Sam and Radford, Alec and Sutskever, Ilya and Amodei, Dario},
 booktitle = {Advances in Neural Information Processing Systems},
 editor = {H. Larochelle and M. Ranzato and R. Hadsell and M.F. Balcan and H. Lin},
 pages = {1877--1901},
 publisher = {Curran Associates, Inc.},
 title = {Language Models are Few-Shot Learners},
 url = {https://proceedings.neurips.cc/paper_files/paper/2020/file/1457c0d6bfcb4967418bfb8ac142f64a-Paper.pdf},
 volume = {33},
 year = {2020}
}

@article{bubeckSparksArtificialGeneral2023,
  author        = {{Bubeck}, Sébastien and {Chandrasekaran}, Varun and {Eldan}, Ronen and {Gehrke}, Johannes and {Horvitz}, Eric and {Kamar}, Ece and {Lee}, Peter and {Lee}, Yin Tat and {Li}, Yuanzhi and {Lundberg}, Scott and {Nori},
Harsha and {Palangi}, Hamid and {Ribeiro}, Marco Tulio and {Zhang}, Yi},
  eprint        = {2303.12712},
  journal       = {arXiv preprints},
  title         = {Sparks of Artificial General Intelligence: Early experiments with GPT-4},
  year          = {2023}
}

@inproceedings{huLoRALowRankAdaptation2021,
  author       = {Edward J. Hu and
                  Yelong Shen and
                  Phillip Wallis and
                  Zeyuan Allen{-}Zhu and
                  Yuanzhi Li and
                  Shean Wang and
                  Lu Wang and
                  Weizhu Chen},
  title        = {LoRA: Low-Rank Adaptation of Large Language Models},
  booktitle    = {The Tenth International Conference on Learning Representations, {ICLR}
                  2022, Virtual Event, April 25-29, 2022},
  publisher    = {OpenReview.net},
  year         = {2022},
  url          = {https://openreview.net/forum?id=nZeVKeeFYf9},
  bibsource    = {dblp computer science bibliography, https://dblp.org}
}

@misc{openaiGPT4TechnicalReport2024,
  title = {{{GPT-4 Technical Report}}},
  author = {OpenAI and Achiam, Josh and Adler, Steven and Agarwal, Sandhini and Ahmad, Lama and Akkaya, Ilge and Aleman, Florencia Leoni and Almeida, Diogo and Altenschmidt, Janko and Altman, Sam and Anadkat, Shyamal and Avila, Red and Babuschkin, Igor and Balaji, Suchir and Balcom, Valerie and Baltescu, Paul and Bao, Haiming and Bavarian, Mohammad and Belgum, Jeff and Bello, Irwan and Berdine, Jake and {Bernadett-Shapiro}, Gabriel and Berner, Christopher and Bogdonoff, Lenny and Boiko, Oleg and Boyd, Madelaine and Brakman, Anna-Luisa and Brockman, Greg and Brooks, Tim and Brundage, Miles and Button, Kevin and Cai, Trevor and Campbell, Rosie and Cann, Andrew and Carey, Brittany and Carlson, Chelsea and Carmichael, Rory and Chan, Brooke and Chang, Che and Chantzis, Fotis and Chen, Derek and Chen, Sully and Chen, Ruby and Chen, Jason and Chen, Mark and Chess, Ben and Cho, Chester and Chu, Casey and Chung, Hyung Won and Cummings, Dave and Currier, Jeremiah and Dai, Yunxing and Decareaux, Cory and Degry, Thomas and Deutsch, Noah and Deville, Damien and Dhar, Arka and Dohan, David and Dowling, Steve and Dunning, Sheila and Ecoffet, Adrien and Eleti, Atty and Eloundou, Tyna and Farhi, David and Fedus, Liam and Felix, Niko and Fishman, Sim{\'o}n Posada and Forte, Juston and Fulford, Isabella and Gao, Leo and Georges, Elie and Gibson, Christian and Goel, Vik and Gogineni, Tarun and Goh, Gabriel and {Gontijo-Lopes}, Rapha and Gordon, Jonathan and Grafstein, Morgan and Gray, Scott and Greene, Ryan and Gross, Joshua and Gu, Shixiang Shane and Guo, Yufei and Hallacy, Chris and Han, Jesse and Harris, Jeff and He, Yuchen and Heaton, Mike and Heidecke, Johannes and Hesse, Chris and Hickey, Alan and Hickey, Wade and Hoeschele, Peter and Houghton, Brandon and Hsu, Kenny and Hu, Shengli and Hu, Xin and Huizinga, Joost and Jain, Shantanu and Jain, Shawn and Jang, Joanne and Jiang, Angela and Jiang, Roger and Jin, Haozhun and Jin, Denny and Jomoto, Shino and Jonn, Billie and Jun, Heewoo and Kaftan, Tomer and Kaiser, {\L}ukasz and Kamali, Ali and Kanitscheider, Ingmar and Keskar, Nitish Shirish and Khan, Tabarak and Kilpatrick, Logan and Kim, Jong Wook and Kim, Christina and Kim, Yongjik and Kirchner, Jan Hendrik and Kiros, Jamie and Knight, Matt and Kokotajlo, Daniel and Kondraciuk, {\L}ukasz and Kondrich, Andrew and Konstantinidis, Aris and Kosic, Kyle and Krueger, Gretchen and Kuo, Vishal and Lampe, Michael and Lan, Ikai and Lee, Teddy and Leike, Jan and Leung, Jade and Levy, Daniel and Li, Chak Ming and Lim, Rachel and Lin, Molly and Lin, Stephanie and Litwin, Mateusz and Lopez, Theresa and Lowe, Ryan and Lue, Patricia and Makanju, Anna and Malfacini, Kim and Manning, Sam and Markov, Todor and Markovski, Yaniv and Martin, Bianca and Mayer, Katie and Mayne, Andrew and McGrew, Bob and McKinney, Scott Mayer and McLeavey, Christine and McMillan, Paul and McNeil, Jake and Medina, David and Mehta, Aalok and Menick, Jacob and Metz, Luke and Mishchenko, Andrey and Mishkin, Pamela and Monaco, Vinnie and Morikawa, Evan and Mossing, Daniel and Mu, Tong and Murati, Mira and Murk, Oleg and M{\'e}ly, David and Nair, Ashvin and Nakano, Reiichiro and Nayak, Rajeev and Neelakantan, Arvind and Ngo, Richard and Noh, Hyeonwoo and Ouyang, Long and O'Keefe, Cullen and Pachocki, Jakub and Paino, Alex and Palermo, Joe and Pantuliano, Ashley and Parascandolo, Giambattista and Parish, Joel and Parparita, Emy and Passos, Alex and Pavlov, Mikhail and Peng, Andrew and Perelman, Adam and Peres, Filipe de Avila Belbute and Petrov, Michael and Pinto, Henrique Ponde de Oliveira and Michael and Pokorny and Pokrass, Michelle and Pong, Vitchyr H. and Powell, Tolly and Power, Alethea and Power, Boris and Proehl, Elizabeth and Puri, Raul and Radford, Alec and Rae, Jack and Ramesh, Aditya and Raymond, Cameron and Real, Francis and Rimbach, Kendra and Ross, Carl and Rotsted, Bob and Roussez, Henri and Ryder, Nick and Saltarelli, Mario and Sanders, Ted and Santurkar, Shibani and Sastry, Girish and Schmidt, Heather and Schnurr, David and Schulman, John and Selsam, Daniel and Sheppard, Kyla and Sherbakov, Toki and Shieh, Jessica and Shoker, Sarah and Shyam, Pranav and Sidor, Szymon and Sigler, Eric and Simens, Maddie and Sitkin, Jordan and Slama, Katarina and Sohl, Ian and Sokolowsky, Benjamin and Song, Yang and Staudacher, Natalie and Such, Felipe Petroski and Summers, Natalie and Sutskever, Ilya and Tang, Jie and Tezak, Nikolas and Thompson, Madeleine B. and Tillet, Phil and Tootoonchian, Amin and Tseng, Elizabeth and Tuggle, Preston and Turley, Nick and Tworek, Jerry and Uribe, Juan Felipe Cer{\'o}n and Vallone, Andrea and Vijayvergiya, Arun and Voss, Chelsea and Wainwright, Carroll and Wang, Justin Jay and Wang, Alvin and Wang, Ben and Ward, Jonathan and Wei, Jason and Weinmann, C. J. and Welihinda, Akila and Welinder, Peter and Weng, Jiayi and Weng, Lilian and Wiethoff, Matt and Willner, Dave and Winter, Clemens and Wolrich, Samuel and Wong, Hannah and Workman, Lauren and Wu, Sherwin and Wu, Jeff and Wu, Michael and Xiao, Kai and Xu, Tao and Yoo, Sarah and Yu, Kevin and Yuan, Qiming and Zaremba, Wojciech and Zellers, Rowan and Zhang, Chong and Zhang, Marvin and Zhao, Shengjia and Zheng, Tianhao and Zhuang, Juntang and Zhuk, William and Zoph, Barret},
  year = {2024},
  month = mar,
  number = {arXiv:2303.08774},
  eprint = {2303.08774},
  primaryclass = {cs},
  publisher = {arXiv},
  doi = {10.48550/arXiv.2303.08774},
  urldate = {2024-08-27},
  archiveprefix = {arXiv}
}

@article{singhalLargeLanguageModels2023,
  title = {Large Language Models Encode Clinical Knowledge},
  author = {Singhal, Karan and Azizi, Shekoofeh and Tu, Tao and Mahdavi, S. Sara and Wei, Jason and Chung, Hyung Won and Scales, Nathan and Tanwani, Ajay and {Cole-Lewis}, Heather and Pfohl, Stephen and Payne, Perry and Seneviratne, Martin and Gamble, Paul and Kelly, Chris and Babiker, Abubakr and Sch{\"a}rli, Nathanael and Chowdhery, Aakanksha and Mansfield, Philip and {Demner-Fushman}, Dina and Ag{\"u}era Y Arcas, Blaise and Webster, Dale and Corrado, Greg S. and Matias, Yossi and Chou, Katherine and Gottweis, Juraj and Tomasev, Nenad and Liu, Yun and Rajkomar, Alvin and Barral, Joelle and Semturs, Christopher and Karthikesalingam, Alan and Natarajan, Vivek},
  year = {2023},
  month = aug,
  journal = {Nature},
  volume = {620},
  number = {7972},
  pages = {172--180},
  issn = {0028-0836, 1476-4687},
  doi = {10.1038/s41586-023-06291-2},
  urldate = {2023-11-21},
  langid = {english}
}

@article{singhalExpertlevelMedicalQuestion2025,
  title = {Toward Expert-Level Medical Question Answering with Large Language Models},
  author = {Singhal, Karan and Tu, Tao and Gottweis, Juraj and Sayres, Rory and Wulczyn, Ellery and Amin, Mohamed and Hou, Le and Clark, Kevin and Pfohl, Stephen R. and {Cole-Lewis}, Heather and Neal, Darlene and Rashid, Qazi Mamunur and Schaekermann, Mike and Wang, Amy and Dash, Dev and Chen, Jonathan H. and Shah, Nigam H. and Lachgar, Sami and Mansfield, Philip Andrew and Prakash, Sushant and Green, Bradley and Dominowska, Ewa and {Ag{\"u}era y Arcas}, Blaise and Toma{\v s}ev, Nenad and Liu, Yun and Wong, Renee and Semturs, Christopher and Mahdavi, S. Sara and Barral, Joelle K. and Webster, Dale R. and Corrado, Greg S. and Matias, Yossi and Azizi, Shekoofeh and Karthikesalingam, Alan and Natarajan, Vivek},
  year = {2025},
  month = mar,
  journal = {Nature Medicine},
  volume = {31},
  number = {3},
  pages = {943--950},
  publisher = {Nature Publishing Group},
  issn = {1546-170X},
  doi = {10.1038/s41591-024-03423-7},
  urldate = {2025-04-08},
  copyright = {2025 The Author(s)},
  langid = {english}
}

@article{vanveenAdaptedLargeLanguage2024,
  title = {Adapted Large Language Models Can Outperform Medical Experts in Clinical Text Summarization},
  author = {Van Veen, Dave and Van Uden, Cara and Blankemeier, Louis and Delbrouck, Jean-Benoit and Aali, Asad and Bluethgen, Christian and Pareek, Anuj and Polacin, Malgorzata and Reis, Eduardo Pontes and Seehofnerov{\'a}, Anna and Rohatgi, Nidhi and Hosamani, Poonam and Collins, William and Ahuja, Neera and Langlotz, Curtis P. and Hom, Jason and Gatidis, Sergios and Pauly, John and Chaudhari, Akshay S.},
  year = {2024},
  month = apr,
  journal = {Nature Medicine},
  volume = {30},
  number = {4},
  pages = {1134--1142},
  publisher = {Nature Publishing Group},
  issn = {1546-170X},
  doi = {10.1038/s41591-024-02855-5},
  urldate = {2024-07-23},
  copyright = {2024 The Author(s), under exclusive licence to Springer Nature America, Inc.},
  langid = {english}
}

@article{alaaCardiovascularDiseaseRisk2019,
  title = {Cardiovascular Disease Risk Prediction Using Automated Machine Learning: {{A}} Prospective Study of 423,604 {{UK Biobank}} Participants},
  shorttitle = {Cardiovascular Disease Risk Prediction Using Automated Machine Learning},
  author = {Alaa, Ahmed M. and Bolton, Thomas and Di Angelantonio, Emanuele and Rudd, James H. F. and Van Der Schaar, Mihaela},
  editor = {{Aalto-Setala}, Katriina},
  year = {2019},
  month = may,
  journal = {PLOS ONE},
  volume = {14},
  number = {5},
  pages = {e0213653},
  issn = {1932-6203},
  doi = {10.1371/journal.pone.0213653},
  urldate = {2024-07-02},
  langid = {english}
}

@article{wengCanMachinelearningImprove2017,
  title = {Can Machine-Learning Improve Cardiovascular Risk Prediction Using Routine Clinical Data?},
  author = {Weng, Stephen F. and Reps, Jenna and Kai, Joe and Garibaldi, Jonathan M. and Qureshi, Nadeem},
  year = {2017},
  month = apr,
  journal = {PLOS ONE},
  volume = {12},
  number = {4},
  pages = {e0174944},
  publisher = {Public Library of Science},
  issn = {1932-6203},
  doi = {10.1371/journal.pone.0174944},
  urldate = {2024-07-20},
  langid = {english}
}

@article{alaaCardiovascularDiseaseRisk2019a,
  title = {Cardiovascular Disease Risk Prediction Using Automated Machine Learning: {{A}} Prospective Study of 423,604 {{UK Biobank}} Participants},
  shorttitle = {Cardiovascular Disease Risk Prediction Using Automated Machine Learning},
  author = {Alaa, Ahmed M. and Bolton, Thomas and Angelantonio, Emanuele Di and Rudd, James H. F. and van der Schaar, Mihaela},
  year = {2019},
  month = may,
  journal = {PLOS ONE},
  volume = {14},
  number = {5},
  pages = {e0213653},
  publisher = {Public Library of Science},
  issn = {1932-6203},
  doi = {10.1371/journal.pone.0213653},
  urldate = {2024-07-20},
  langid = {english}
}

@article{singhalLargeLanguageModels2023a,
  title = {Large Language Models Encode Clinical Knowledge},
  author = {Singhal, Karan and Azizi, Shekoofeh and Tu, Tao and Mahdavi, S. Sara and Wei, Jason and Chung, Hyung Won and Scales, Nathan and Tanwani, Ajay and {Cole-Lewis}, Heather and Pfohl, Stephen and Payne, Perry and Seneviratne, Martin and Gamble, Paul and Kelly, Chris and Babiker, Abubakr and Sch{\"a}rli, Nathanael and Chowdhery, Aakanksha and Mansfield, Philip and {Demner-Fushman}, Dina and {Ag{\"u}era y Arcas}, Blaise and Webster, Dale and Corrado, Greg S. and Matias, Yossi and Chou, Katherine and Gottweis, Juraj and Tomasev, Nenad and Liu, Yun and Rajkomar, Alvin and Barral, Joelle and Semturs, Christopher and Karthikesalingam, Alan and Natarajan, Vivek},
  year = {2023},
  month = aug,
  journal = {Nature},
  volume = {620},
  number = {7972},
  pages = {172--180},
  publisher = {Nature Publishing Group},
  issn = {1476-4687},
  doi = {10.1038/s41586-023-06291-2},
  urldate = {2024-07-21},
  copyright = {2023 The Author(s)},
  langid = {english}
}

@inproceedings{hegselmannTabllmFewshotClassification2023,
  title = {Tabllm: {{Few-shot}} Classification of Tabular Data with Large Language Models},
  shorttitle = {Tabllm},
  booktitle = {International {{Conference}} on {{Artificial Intelligence}} and {{Statistics}}},
  author = {Hegselmann, Stefan and Buendia, Alejandro and Lang, Hunter and Agrawal, Monica and Jiang, Xiaoyi and Sontag, David},
  year = {2023},
  pages = {5549--5581},
  publisher = {PMLR},
  urldate = {2024-07-21}
}

@article{hanEvaluationGPT410year2024a,
  title = {Evaluation of {{GPT-4}} for 10-Year Cardiovascular Risk Prediction: {{Insights}} from the {{UK Biobank}} and {{KoGES}} Data},
  shorttitle = {Evaluation of {{GPT-4}} for 10-Year Cardiovascular Risk Prediction},
  author = {Han, Changho and Kim, Dong Won and Kim, Songsoo and Chan You, Seng and Park, Jin Young and Bae, SungA and Yoon, Dukyong},
  year = {2024},
  month = feb,
  journal = {iScience},
  volume = {27},
  number = {2},
  pages = {109022},
  issn = {2589-0042},
  doi = {10.1016/j.isci.2024.109022},
  langid = {english},
  pmcid = {PMC10865411},
  pmid = {38357664}
}

@misc{belyaevaMultimodalLLMsHealth2023a,
  title = {Multimodal {{LLMs}} for Health Grounded in Individual-Specific Data},
  author = {Belyaeva, Anastasiya and Cosentino, Justin and Hormozdiari, Farhad and Eswaran, Krish and Shetty, Shravya and Corrado, Greg and Carroll, Andrew and McLean, Cory Y. and Furlotte, Nicholas A.},
  year = {2023},
  month = jul,
  number = {arXiv:2307.09018},
  eprint = {2307.09018},
  primaryclass = {cs, q-bio},
  publisher = {arXiv},
  doi = {10.48550/arXiv.2307.09018},
  urldate = {2024-07-21},
  archiveprefix = {arXiv}
}

@article{moorFoundationModelsGeneralist2023,
  title = {Foundation Models for Generalist Medical Artificial Intelligence},
  author = {Moor, Michael and Banerjee, Oishi and Abad, Zahra Shakeri Hossein and Krumholz, Harlan M. and Leskovec, Jure and Topol, Eric J. and Rajpurkar, Pranav},
  year = {2023},
  month = apr,
  journal = {Nature},
  volume = {616},
  number = {7956},
  pages = {259--265},
  publisher = {Nature Publishing Group},
  issn = {1476-4687},
  doi = {10.1038/s41586-023-05881-4},
  urldate = {2024-07-24},
  copyright = {2023 Springer Nature Limited},
  langid = {english}
}

@misc{CardiovascularDiseasesCVDs,
  title = {Cardiovascular Diseases ({{CVDs}})},
  urldate = {2024-07-24},
  howpublished = {https://www.who.int/news-room/fact-sheets/detail/cardiovascular-diseases-(cvds)},
  langid = {english}
}

@article{khanDevelopmentValidationAmerican2024,
  title = {Development and {{Validation}} of the {{American Heart Association}}'s {{PREVENT Equations}}},
  author = {Khan, Sadiya S. and Matsushita, Kunihiro and Sang, Yingying and Ballew, Shoshana H. and Grams, Morgan E. and Surapaneni, Aditya and Blaha, Michael J. and Carson, April P. and Chang, Alexander R. and Ciemins, Elizabeth and Go, Alan S. and Gutierrez, Orlando M. and Hwang, Shih-Jen and Jassal, Simerjot K. and Kovesdy, Csaba P. and {Lloyd-Jones}, Donald M. and Shlipak, Michael G. and Palaniappan, Latha P. and Sperling, Laurence and Virani, Salim S. and Tuttle, Katherine and Neeland, Ian J. and Chow, Sheryl L. and Rangaswami, Janani and Pencina, Michael J. and Ndumele, Chiadi E. and Coresh, Josef and {for the Chronic Kidney Disease Prognosis Consortium and the American Heart Association Cardiovascular-Kidney-Metabolic Science Advisory Group}},
  year = {2024},
  month = feb,
  journal = {Circulation},
  volume = {149},
  number = {6},
  pages = {430--449},
  issn = {0009-7322, 1524-4539},
  doi = {10.1161/CIRCULATIONAHA.123.067626},
  urldate = {2024-07-26},
  langid = {english}
}

@article{dagostinoGeneralCardiovascularRisk2008,
  title = {General {{Cardiovascular Risk Profile}} for {{Use}} in {{Primary Care}}: {{The Framingham Heart Study}}},
  shorttitle = {General {{Cardiovascular Risk Profile}} for {{Use}} in {{Primary Care}}},
  author = {D'Agostino, Ralph B. and Vasan, Ramachandran S. and Pencina, Michael J. and Wolf, Philip A. and Cobain, Mark and Massaro, Joseph M. and Kannel, William B.},
  year = {2008},
  month = feb,
  journal = {Circulation},
  volume = {117},
  number = {6},
  pages = {743--753},
  issn = {0009-7322, 1524-4539},
  doi = {10.1161/CIRCULATIONAHA.107.699579},
  urldate = {2024-07-26},
  langid = {english}
}

@article{arnett2019ACCAHA2019,
  title = {2019 {{ACC}}/{{AHA Guideline}} on the {{Primary Prevention}} of {{Cardiovascular Disease}}: {{A Report}} of the {{American College}} of {{Cardiology}}/{{American Heart Association Task Force}} on {{Clinical Practice Guidelines}}},
  shorttitle = {2019 {{ACC}}/{{AHA Guideline}} on the {{Primary Prevention}} of {{Cardiovascular Disease}}},
  author = {Arnett, Donna K. and Blumenthal, Roger S. and Albert, Michelle A. and Buroker, Andrew B. and Goldberger, Zachary D. and Hahn, Ellen J. and Himmelfarb, Cheryl Dennison and Khera, Amit and {Lloyd-Jones}, Donald and McEvoy, J. William and Michos, Erin D. and Miedema, Michael D. and Mu{\~n}oz, Daniel and Smith, Sidney C. and Virani, Salim S. and Williams, Kim A. and Yeboah, Joseph and Ziaeian, Boback},
  year = {2019},
  month = sep,
  journal = {Circulation},
  volume = {140},
  number = {11},
  issn = {0009-7322, 1524-4539},
  doi = {10.1161/CIR.0000000000000678},
  urldate = {2024-07-26},
  langid = {english}
}

@article{dingParameterefficientFinetuningLargescale2023,
  title = {Parameter-Efficient Fine-Tuning of Large-Scale Pre-Trained Language Models},
  author = {Ding, Ning and Qin, Yujia and Yang, Guang and Wei, Fuchao and Yang, Zonghan and Su, Yusheng and Hu, Shengding and Chen, Yulin and Chan, Chi-Min and Chen, Weize and Yi, Jing and Zhao, Weilin and Wang, Xiaozhi and Liu, Zhiyuan and Zheng, Hai-Tao and Chen, Jianfei and Liu, Yang and Tang, Jie and Li, Juanzi and Sun, Maosong},
  year = {2023},
  month = mar,
  journal = {Nature Machine Intelligence},
  volume = {5},
  number = {3},
  pages = {220--235},
  publisher = {Nature Publishing Group},
  issn = {2522-5839},
  doi = {10.1038/s42256-023-00626-4},
  urldate = {2024-07-26},
  copyright = {2023 The Author(s)},
  langid = {english}
}

@misc{jiangMistral7B2023,
  title = {Mistral {{7B}}},
  author = {Jiang, Albert Q. and Sablayrolles, Alexandre and Mensch, Arthur and Bamford, Chris and Chaplot, Devendra Singh and de las Casas, Diego and Bressand, Florian and Lengyel, Gianna and Lample, Guillaume and Saulnier, Lucile and Lavaud, L{\'e}lio Renard and Lachaux, Marie-Anne and Stock, Pierre and Scao, Teven Le and Lavril, Thibaut and Wang, Thomas and Lacroix, Timoth{\'e}e and Sayed, William El},
  year = {2023},
  month = oct,
  number = {arXiv:2310.06825},
  eprint = {2310.06825},
  primaryclass = {cs},
  publisher = {arXiv},
  doi = {10.48550/arXiv.2310.06825},
  urldate = {2025-01-14},
  archiveprefix = {arXiv}
}

@misc{grattafioriLlamaHerdModels2024,
  title = {The {{Llama}} 3 {{Herd}} of {{Models}}},
  author = {Grattafiori, Aaron and Dubey, Abhimanyu and Jauhri, Abhinav and Pandey, Abhinav and Kadian, Abhishek and {Al-Dahle}, Ahmad and Letman, Aiesha and Mathur, Akhil and Schelten, Alan and Vaughan, Alex and Yang, Amy and Fan, Angela and Goyal, Anirudh and Hartshorn, Anthony and Yang, Aobo and Mitra, Archi and Sravankumar, Archie and Korenev, Artem and Hinsvark, Arthur and Rao, Arun and Zhang, Aston and Rodriguez, Aurelien and Gregerson, Austen and Spataru, Ava and Roziere, Baptiste and Biron, Bethany and Tang, Binh and Chern, Bobbie and Caucheteux, Charlotte and Nayak, Chaya and Bi, Chloe and Marra, Chris and McConnell, Chris and Keller, Christian and Touret, Christophe and Wu, Chunyang and Wong, Corinne and Ferrer, Cristian Canton and Nikolaidis, Cyrus and Allonsius, Damien and Song, Daniel and Pintz, Danielle and Livshits, Danny and Wyatt, Danny and Esiobu, David and Choudhary, Dhruv and Mahajan, Dhruv and {Garcia-Olano}, Diego and Perino, Diego and Hupkes, Dieuwke and Lakomkin, Egor and AlBadawy, Ehab and Lobanova, Elina and Dinan, Emily and Smith, Eric Michael and Radenovic, Filip and Guzm{\'a}n, Francisco and Zhang, Frank and Synnaeve, Gabriel and Lee, Gabrielle and Anderson, Georgia Lewis and Thattai, Govind and Nail, Graeme and Mialon, Gregoire and Pang, Guan and Cucurell, Guillem and Nguyen, Hailey and Korevaar, Hannah and Xu, Hu and Touvron, Hugo and Zarov, Iliyan and Ibarra, Imanol Arrieta and Kloumann, Isabel and Misra, Ishan and Evtimov, Ivan and Zhang, Jack and Copet, Jade and Lee, Jaewon and Geffert, Jan and Vranes, Jana and Park, Jason and Mahadeokar, Jay and Shah, Jeet and van der Linde, Jelmer and Billock, Jennifer and Hong, Jenny and Lee, Jenya and Fu, Jeremy and Chi, Jianfeng and Huang, Jianyu and Liu, Jiawen and Wang, Jie and Yu, Jiecao and Bitton, Joanna and Spisak, Joe and Park, Jongsoo and Rocca, Joseph and Johnstun, Joshua and Saxe, Joshua and Jia, Junteng and Alwala, Kalyan Vasuden and Prasad, Karthik and Upasani, Kartikeya and Plawiak, Kate and Li, Ke and Heafield, Kenneth and Stone, Kevin and {El-Arini}, Khalid and Iyer, Krithika and Malik, Kshitiz and Chiu, Kuenley and Bhalla, Kunal and Lakhotia, Kushal and {Rantala-Yeary}, Lauren and van der Maaten, Laurens and Chen, Lawrence and Tan, Liang and Jenkins, Liz and Martin, Louis and Madaan, Lovish and Malo, Lubo and Blecher, Lukas and Landzaat, Lukas and de Oliveira, Luke and Muzzi, Madeline and Pasupuleti, Mahesh and Singh, Mannat and Paluri, Manohar and Kardas, Marcin and Tsimpoukelli, Maria and Oldham, Mathew and Rita, Mathieu and Pavlova, Maya and Kambadur, Melanie and Lewis, Mike and Si, Min and Singh, Mitesh Kumar and Hassan, Mona and Goyal, Naman and Torabi, Narjes and Bashlykov, Nikolay and Bogoychev, Nikolay and Chatterji, Niladri and Zhang, Ning and Duchenne, Olivier and {\c C}elebi, Onur and Alrassy, Patrick and Zhang, Pengchuan and Li, Pengwei and Vasic, Petar and Weng, Peter and Bhargava, Prajjwal and Dubal, Pratik and Krishnan, Praveen and Koura, Punit Singh and Xu, Puxin and He, Qing and Dong, Qingxiao and Srinivasan, Ragavan and Ganapathy, Raj and Calderer, Ramon and Cabral, Ricardo Silveira and Stojnic, Robert and Raileanu, Roberta and Maheswari, Rohan and Girdhar, Rohit and Patel, Rohit and Sauvestre, Romain and Polidoro, Ronnie and Sumbaly, Roshan and Taylor, Ross and Silva, Ruan and Hou, Rui and Wang, Rui and Hosseini, Saghar and Chennabasappa, Sahana and Singh, Sanjay and Bell, Sean and Kim, Seohyun Sonia and Edunov, Sergey and Nie, Shaoliang and Narang, Sharan and Raparthy, Sharath and Shen, Sheng and Wan, Shengye and Bhosale, Shruti and Zhang, Shun and Vandenhende, Simon and Batra, Soumya and Whitman, Spencer and Sootla, Sten and Collot, Stephane and Gururangan, Suchin and Borodinsky, Sydney and Herman, Tamar and Fowler, Tara and Sheasha, Tarek and Georgiou, Thomas and Scialom, Thomas and Speckbacher, Tobias and Mihaylov, Todor and Xiao, Tong and Karn, Ujjwal and Goswami, Vedanuj and Gupta, Vibhor and Ramanathan, Vignesh and Kerkez, Viktor and Gonguet, Vincent and Do, Virginie and Vogeti, Vish and Albiero, V{\'i}tor and Petrovic, Vladan and Chu, Weiwei and Xiong, Wenhan and Fu, Wenyin and Meers, Whitney and Martinet, Xavier and Wang, Xiaodong and Wang, Xiaofang and Tan, Xiaoqing Ellen and Xia, Xide and Xie, Xinfeng and Jia, Xuchao and Wang, Xuewei and Goldschlag, Yaelle and Gaur, Yashesh and Babaei, Yasmine and Wen, Yi and Song, Yiwen and Zhang, Yuchen and Li, Yue and Mao, Yuning and Coudert, Zacharie Delpierre and Yan, Zheng and Chen, Zhengxing and Papakipos, Zoe and Singh, Aaditya and Srivastava, Aayushi and Jain, Abha and Kelsey, Adam and Shajnfeld, Adam and Gangidi, Adithya and Victoria, Adolfo and Goldstand, Ahuva and Menon, Ajay and Sharma, Ajay and Boesenberg, Alex and Baevski, Alexei and Feinstein, Allie and Kallet, Amanda and Sangani, Amit and Teo, Amos and Yunus, Anam and Lupu, Andrei and Alvarado, Andres and Caples, Andrew and Gu, Andrew and Ho, Andrew and Poulton, Andrew and Ryan, Andrew and Ramchandani, Ankit and Dong, Annie and Franco, Annie and Goyal, Anuj and Saraf, Aparajita and Chowdhury, Arkabandhu and Gabriel, Ashley and Bharambe, Ashwin and Eisenman, Assaf and Yazdan, Azadeh and James, Beau and Maurer, Ben and Leonhardi, Benjamin and Huang, Bernie and Loyd, Beth and Paola, Beto De and Paranjape, Bhargavi and Liu, Bing and Wu, Bo and Ni, Boyu and Hancock, Braden and Wasti, Bram and Spence, Brandon and Stojkovic, Brani and Gamido, Brian and Montalvo, Britt and Parker, Carl and Burton, Carly and Mejia, Catalina and Liu, Ce and Wang, Changhan and Kim, Changkyu and Zhou, Chao and Hu, Chester and Chu, Ching-Hsiang and Cai, Chris and Tindal, Chris and Feichtenhofer, Christoph and Gao, Cynthia and Civin, Damon and Beaty, Dana and Kreymer, Daniel and Li, Daniel and Adkins, David and Xu, David and Testuggine, Davide and David, Delia and Parikh, Devi and Liskovich, Diana and Foss, Didem and Wang, Dingkang and Le, Duc and Holland, Dustin and Dowling, Edward and Jamil, Eissa and Montgomery, Elaine and Presani, Eleonora and Hahn, Emily and Wood, Emily and Le, Eric-Tuan and Brinkman, Erik and Arcaute, Esteban and Dunbar, Evan and Smothers, Evan and Sun, Fei and Kreuk, Felix and Tian, Feng and Kokkinos, Filippos and Ozgenel, Firat and Caggioni, Francesco and Kanayet, Frank and Seide, Frank and Florez, Gabriela Medina and Schwarz, Gabriella and Badeer, Gada and Swee, Georgia and Halpern, Gil and Herman, Grant and Sizov, Grigory and Guangyi and Zhang and Lakshminarayanan, Guna and Inan, Hakan and Shojanazeri, Hamid and Zou, Han and Wang, Hannah and Zha, Hanwen and Habeeb, Haroun and Rudolph, Harrison and Suk, Helen and Aspegren, Henry and Goldman, Hunter and Zhan, Hongyuan and Damlaj, Ibrahim and Molybog, Igor and Tufanov, Igor and Leontiadis, Ilias and Veliche, Irina-Elena and Gat, Itai and Weissman, Jake and Geboski, James and Kohli, James and Lam, Janice and Asher, Japhet and Gaya, Jean-Baptiste and Marcus, Jeff and Tang, Jeff and Chan, Jennifer and Zhen, Jenny and Reizenstein, Jeremy and Teboul, Jeremy and Zhong, Jessica and Jin, Jian and Yang, Jingyi and Cummings, Joe and Carvill, Jon and Shepard, Jon and McPhie, Jonathan and Torres, Jonathan and Ginsburg, Josh and Wang, Junjie and Wu, Kai and U, Kam Hou and Saxena, Karan and Khandelwal, Kartikay and Zand, Katayoun and Matosich, Kathy and Veeraraghavan, Kaushik and Michelena, Kelly and Li, Keqian and Jagadeesh, Kiran and Huang, Kun and Chawla, Kunal and Huang, Kyle and Chen, Lailin and Garg, Lakshya and A, Lavender and Silva, Leandro and Bell, Lee and Zhang, Lei and Guo, Liangpeng and Yu, Licheng and Moshkovich, Liron and Wehrstedt, Luca and Khabsa, Madian and Avalani, Manav and Bhatt, Manish and Mankus, Martynas and Hasson, Matan and Lennie, Matthew and Reso, Matthias and Groshev, Maxim and Naumov, Maxim and Lathi, Maya and Keneally, Meghan and Liu, Miao and Seltzer, Michael L. and Valko, Michal and Restrepo, Michelle and Patel, Mihir and Vyatskov, Mik and Samvelyan, Mikayel and Clark, Mike and Macey, Mike and Wang, Mike and Hermoso, Miquel Jubert and Metanat, Mo and Rastegari, Mohammad and Bansal, Munish and Santhanam, Nandhini and Parks, Natascha and White, Natasha and Bawa, Navyata and Singhal, Nayan and Egebo, Nick and Usunier, Nicolas and Mehta, Nikhil and Laptev, Nikolay Pavlovich and Dong, Ning and Cheng, Norman and Chernoguz, Oleg and Hart, Olivia and Salpekar, Omkar and Kalinli, Ozlem and Kent, Parkin and Parekh, Parth and Saab, Paul and Balaji, Pavan and Rittner, Pedro and Bontrager, Philip and Roux, Pierre and Dollar, Piotr and Zvyagina, Polina and Ratanchandani, Prashant and Yuvraj, Pritish and Liang, Qian and Alao, Rachad and Rodriguez, Rachel and Ayub, Rafi and Murthy, Raghotham and Nayani, Raghu and Mitra, Rahul and Parthasarathy, Rangaprabhu and Li, Raymond and Hogan, Rebekkah and Battey, Robin and Wang, Rocky and Howes, Russ and Rinott, Ruty and Mehta, Sachin and Siby, Sachin and Bondu, Sai Jayesh and Datta, Samyak and Chugh, Sara and Hunt, Sara and Dhillon, Sargun and Sidorov, Sasha and Pan, Satadru and Mahajan, Saurabh and Verma, Saurabh and Yamamoto, Seiji and Ramaswamy, Sharadh and Lindsay, Shaun and Lindsay, Shaun and Feng, Sheng and Lin, Shenghao and Zha, Shengxin Cindy and Patil, Shishir and Shankar, Shiva and Zhang, Shuqiang and Zhang, Shuqiang and Wang, Sinong and Agarwal, Sneha and Sajuyigbe, Soji and Chintala, Soumith and Max, Stephanie and Chen, Stephen and Kehoe, Steve and Satterfield, Steve and Govindaprasad, Sudarshan and Gupta, Sumit and Deng, Summer and Cho, Sungmin and Virk, Sunny and Subramanian, Suraj and Choudhury, Sy and Goldman, Sydney and Remez, Tal and Glaser, Tamar and Best, Tamara and Koehler, Thilo and Robinson, Thomas and Li, Tianhe and Zhang, Tianjun and Matthews, Tim and Chou, Timothy and Shaked, Tzook and Vontimitta, Varun and Ajayi, Victoria and Montanez, Victoria and Mohan, Vijai and Kumar, Vinay Satish and Mangla, Vishal and Ionescu, Vlad and Poenaru, Vlad and Mihailescu, Vlad Tiberiu and Ivanov, Vladimir and Li, Wei and Wang, Wenchen and Jiang, Wenwen and Bouaziz, Wes and Constable, Will and Tang, Xiaocheng and Wu, Xiaojian and Wang, Xiaolan and Wu, Xilun and Gao, Xinbo and Kleinman, Yaniv and Chen, Yanjun and Hu, Ye and Jia, Ye and Qi, Ye and Li, Yenda and Zhang, Yilin and Zhang, Ying and Adi, Yossi and Nam, Youngjin and Yu and Wang and Zhao, Yu and Hao, Yuchen and Qian, Yundi and Li, Yunlu and He, Yuzi and Rait, Zach and DeVito, Zachary and Rosnbrick, Zef and Wen, Zhaoduo and Yang, Zhenyu and Zhao, Zhiwei and Ma, Zhiyu},
  year = {2024},
  month = nov,
  number = {arXiv:2407.21783},
  eprint = {2407.21783},
  primaryclass = {cs},
  publisher = {arXiv},
  doi = {10.48550/arXiv.2407.21783},
  urldate = {2025-01-14},
  archiveprefix = {arXiv}
}

@misc{abdinPhi3TechnicalReport2024,
  title = {Phi-3 {{Technical Report}}: {{A Highly Capable Language Model Locally}} on {{Your Phone}}},
  shorttitle = {Phi-3 {{Technical Report}}},
  author = {Abdin, Marah and Aneja, Jyoti and Awadalla, Hany and Awadallah, Ahmed and Awan, Ammar Ahmad and Bach, Nguyen and Bahree, Amit and Bakhtiari, Arash and Bao, Jianmin and Behl, Harkirat and Benhaim, Alon and Bilenko, Misha and Bjorck, Johan and Bubeck, S{\'e}bastien and Cai, Martin and Cai, Qin and Chaudhary, Vishrav and Chen, Dong and Chen, Dongdong and Chen, Weizhu and Chen, Yen-Chun and Chen, Yi-Ling and Cheng, Hao and Chopra, Parul and Dai, Xiyang and Dixon, Matthew and Eldan, Ronen and Fragoso, Victor and Gao, Jianfeng and Gao, Mei and Gao, Min and Garg, Amit and Giorno, Allie Del and Goswami, Abhishek and Gunasekar, Suriya and Haider, Emman and Hao, Junheng and Hewett, Russell J. and Hu, Wenxiang and Huynh, Jamie and Iter, Dan and Jacobs, Sam Ade and Javaheripi, Mojan and Jin, Xin and Karampatziakis, Nikos and Kauffmann, Piero and Khademi, Mahoud and Kim, Dongwoo and Kim, Young Jin and Kurilenko, Lev and Lee, James R. and Lee, Yin Tat and Li, Yuanzhi and Li, Yunsheng and Liang, Chen and Liden, Lars and Lin, Xihui and Lin, Zeqi and Liu, Ce and Liu, Liyuan and Liu, Mengchen and Liu, Weishung and Liu, Xiaodong and Luo, Chong and Madan, Piyush and Mahmoudzadeh, Ali and Majercak, David and Mazzola, Matt and Mendes, Caio C{\'e}sar Teodoro and Mitra, Arindam and Modi, Hardik and Nguyen, Anh and Norick, Brandon and Patra, Barun and {Perez-Becker}, Daniel and Portet, Thomas and Pryzant, Reid and Qin, Heyang and Radmilac, Marko and Ren, Liliang and de Rosa, Gustavo and Rosset, Corby and Roy, Sambudha and Ruwase, Olatunji and Saarikivi, Olli and Saied, Amin and Salim, Adil and Santacroce, Michael and Shah, Shital and Shang, Ning and Sharma, Hiteshi and Shen, Yelong and Shukla, Swadheen and Song, Xia and Tanaka, Masahiro and Tupini, Andrea and Vaddamanu, Praneetha and Wang, Chunyu and Wang, Guanhua and Wang, Lijuan and Wang, Shuohang and Wang, Xin and Wang, Yu and Ward, Rachel and Wen, Wen and Witte, Philipp and Wu, Haiping and Wu, Xiaoxia and Wyatt, Michael and Xiao, Bin and Xu, Can and Xu, Jiahang and Xu, Weijian and Xue, Jilong and Yadav, Sonali and Yang, Fan and Yang, Jianwei and Yang, Yifan and Yang, Ziyi and Yu, Donghan and Yuan, Lu and Zhang, Chenruidong and Zhang, Cyril and Zhang, Jianwen and Zhang, Li Lyna and Zhang, Yi and Zhang, Yue and Zhang, Yunan and Zhou, Xiren},
  year = {2024},
  month = aug,
  number = {arXiv:2404.14219},
  eprint = {2404.14219},
  primaryclass = {cs},
  publisher = {arXiv},
  doi = {10.48550/arXiv.2404.14219},
  urldate = {2025-01-14},
  archiveprefix = {arXiv}
}

@misc{teamGemmaImprovingOpen2024,
  title = {Gemma 2: {{Improving Open Language Models}} at a {{Practical Size}}},
  shorttitle = {Gemma 2},
  author = {Team, Gemma and Riviere, Morgane and Pathak, Shreya and Sessa, Pier Giuseppe and Hardin, Cassidy and Bhupatiraju, Surya and Hussenot, L{\'e}onard and Mesnard, Thomas and Shahriari, Bobak and Ram{\'e}, Alexandre and Ferret, Johan and Liu, Peter and Tafti, Pouya and Friesen, Abe and Casbon, Michelle and Ramos, Sabela and Kumar, Ravin and Lan, Charline Le and Jerome, Sammy and Tsitsulin, Anton and Vieillard, Nino and Stanczyk, Piotr and Girgin, Sertan and Momchev, Nikola and Hoffman, Matt and Thakoor, Shantanu and Grill, Jean-Bastien and Neyshabur, Behnam and Bachem, Olivier and Walton, Alanna and Severyn, Aliaksei and Parrish, Alicia and Ahmad, Aliya and Hutchison, Allen and Abdagic, Alvin and Carl, Amanda and Shen, Amy and Brock, Andy and Coenen, Andy and Laforge, Anthony and Paterson, Antonia and Bastian, Ben and Piot, Bilal and Wu, Bo and Royal, Brandon and Chen, Charlie and Kumar, Chintu and Perry, Chris and Welty, Chris and {Choquette-Choo}, Christopher A. and Sinopalnikov, Danila and Weinberger, David and Vijaykumar, Dimple and Rogozi{\'n}ska, Dominika and Herbison, Dustin and Bandy, Elisa and Wang, Emma and Noland, Eric and Moreira, Erica and Senter, Evan and Eltyshev, Evgenii and Visin, Francesco and Rasskin, Gabriel and Wei, Gary and Cameron, Glenn and Martins, Gus and Hashemi, Hadi and {Klimczak-Pluci{\'n}ska}, Hanna and Batra, Harleen and Dhand, Harsh and Nardini, Ivan and Mein, Jacinda and Zhou, Jack and Svensson, James and Stanway, Jeff and Chan, Jetha and Zhou, Jin Peng and Carrasqueira, Joana and Iljazi, Joana and Becker, Jocelyn and Fernandez, Joe and van Amersfoort, Joost and Gordon, Josh and Lipschultz, Josh and Newlan, Josh and Ji, Ju-yeong and Mohamed, Kareem and Badola, Kartikeya and Black, Kat and Millican, Katie and McDonell, Keelin and Nguyen, Kelvin and Sodhia, Kiranbir and Greene, Kish and Sjoesund, Lars Lowe and Usui, Lauren and Sifre, Laurent and Heuermann, Lena and Lago, Leticia and McNealus, Lilly and Soares, Livio Baldini and Kilpatrick, Logan and Dixon, Lucas and Martins, Luciano and Reid, Machel and Singh, Manvinder and Iverson, Mark and G{\"o}rner, Martin and Velloso, Mat and Wirth, Mateo and Davidow, Matt and Miller, Matt and Rahtz, Matthew and Watson, Matthew and Risdal, Meg and Kazemi, Mehran and Moynihan, Michael and Zhang, Ming and Kahng, Minsuk and Park, Minwoo and Rahman, Mofi and Khatwani, Mohit and Dao, Natalie and Bardoliwalla, Nenshad and Devanathan, Nesh and Dumai, Neta and Chauhan, Nilay and Wahltinez, Oscar and Botarda, Pankil and Barnes, Parker and Barham, Paul and Michel, Paul and Jin, Pengchong and Georgiev, Petko and Culliton, Phil and Kuppala, Pradeep and Comanescu, Ramona and Merhej, Ramona and Jana, Reena and Rokni, Reza Ardeshir and Agarwal, Rishabh and Mullins, Ryan and Saadat, Samaneh and Carthy, Sara Mc and Cogan, Sarah and Perrin, Sarah and Arnold, S{\'e}bastien M. R. and Krause, Sebastian and Dai, Shengyang and Garg, Shruti and Sheth, Shruti and Ronstrom, Sue and Chan, Susan and Jordan, Timothy and Yu, Ting and Eccles, Tom and Hennigan, Tom and Kocisky, Tomas and Doshi, Tulsee and Jain, Vihan and Yadav, Vikas and Meshram, Vilobh and Dharmadhikari, Vishal and Barkley, Warren and Wei, Wei and Ye, Wenming and Han, Woohyun and Kwon, Woosuk and Xu, Xiang and Shen, Zhe and Gong, Zhitao and Wei, Zichuan and Cotruta, Victor and Kirk, Phoebe and Rao, Anand and Giang, Minh and Peran, Ludovic and Warkentin, Tris and Collins, Eli and Barral, Joelle and Ghahramani, Zoubin and Hadsell, Raia and Sculley, D. and Banks, Jeanine and Dragan, Anca and Petrov, Slav and Vinyals, Oriol and Dean, Jeff and Hassabis, Demis and Kavukcuoglu, Koray and Farabet, Clement and Buchatskaya, Elena and Borgeaud, Sebastian and Fiedel, Noah and Joulin, Armand and Kenealy, Kathleen and Dadashi, Robert and Andreev, Alek},
  year = {2024},
  month = oct,
  number = {arXiv:2408.00118},
  eprint = {2408.00118},
  primaryclass = {cs},
  publisher = {arXiv},
  doi = {10.48550/arXiv.2408.00118},
  urldate = {2025-01-14},
  archiveprefix = {arXiv}
}

@article{score2workinggroupandesccardiovascularriskcollaborationSCORE2RiskPrediction2021,
  title = {{{SCORE2}} Risk Prediction Algorithms: New Models to Estimate 10-Year Risk of Cardiovascular Disease in {{Europe}}},
  shorttitle = {{{SCORE2}} Risk Prediction Algorithms},
  author = {{SCORE2 working group and ESC Cardiovascular risk collaboration}},
  year = {2021},
  month = jul,
  journal = {European Heart Journal},
  volume = {42},
  number = {25},
  pages = {2439--2454},
  issn = {0195-668X},
  doi = {10.1093/eurheartj/ehab309},
  urldate = {2025-01-14}
}

@article{hippisley-coxDerivationValidationQRISK2007,
  title = {Derivation and Validation of {{QRISK}}, a New Cardiovascular Disease Risk Score for the {{United Kingdom}}: Prospective Open Cohort Study},
  shorttitle = {Derivation and Validation of {{QRISK}}, a New Cardiovascular Disease Risk Score for the {{United Kingdom}}},
  author = {{Hippisley-Cox}, Julia and Coupland, Carol and Vinogradova, Yana and Robson, John and May, Margaret and Brindle, Peter},
  year = {2007},
  month = jul,
  journal = {BMJ : British Medical Journal},
  volume = {335},
  number = {7611},
  pages = {136},
  issn = {0959-8138},
  doi = {10.1136/bmj.39261.471806.55},
  urldate = {2025-01-14},
  pmcid = {PMC1925200},
  pmid = {17615182}
}

@article{rothGlobalBurdenCardiovascular2020,
  title = {The {{Global Burden}} of {{Cardiovascular Diseases}} and {{Risks}}: {{A Compass}} for {{Global Action}}},
  shorttitle = {The {{Global Burden}} of {{Cardiovascular Diseases}} and {{Risks}}},
  author = {Roth, Gregory A. and Mensah, George A. and Fuster, Valentin},
  year = {2020},
  month = dec,
  journal = {Journal of the American College of Cardiology},
  volume = {76},
  number = {25},
  pages = {2980--2981},
  issn = {1558-3597},
  doi = {10.1016/j.jacc.2020.11.021},
  langid = {english},
  pmid = {33309174}
}

@article{khanNovelPredictionEquations2023,
  title = {Novel {{Prediction Equations}} for {{Absolute Risk Assessment}} of {{Total Cardiovascular Disease Incorporating Cardiovascular-Kidney-Metabolic Health}}: {{A Scientific Statement From}} the {{American Heart Association}}},
  shorttitle = {Novel {{Prediction Equations}} for {{Absolute Risk Assessment}} of {{Total Cardiovascular Disease Incorporating Cardiovascular-Kidney-Metabolic Health}}},
  author = {Khan, Sadiya S. and Coresh, Josef and Pencina, Michael J. and Ndumele, Chiadi E. and Rangaswami, Janani and Chow, Sheryl L. and Palaniappan, Latha P. and Sperling, Laurence S. and Virani, Salim S. and Ho, Jennifer E. and Neeland, Ian J. and Tuttle, Katherine R. and Rajgopal Singh, Radhika and Elkind, Mitchell S. V. and {Lloyd-Jones}, Donald M. and {American Heart Association}},
  year = {2023},
  month = dec,
  journal = {Circulation},
  volume = {148},
  number = {24},
  pages = {1982--2004},
  issn = {1524-4539},
  doi = {10.1161/CIR.0000000000001191},
  langid = {english},
  pmid = {37947094}
}

@misc{OverviewCardiovascularDisease2023,
  title = {Overview {\textbar} {{Cardiovascular}} Disease: Risk Assessment and Reduction, Including Lipid Modification {\textbar} {{Guidance}} {\textbar} {{NICE}}},
  shorttitle = {Overview {\textbar} {{Cardiovascular}} Disease},
  year = {2023},
  month = dec,
  publisher = {NICE},
  urldate = {2025-03-06},
  howpublished = {https://www.nice.org.uk/guidance/ng238},
  langid = {english}
}

@article{visseren2021ESCGuidelines2021,
  title = {2021 {{ESC Guidelines}} on Cardiovascular Disease Prevention in Clinical Practice: {{Developed}} by the {{Task Force}} for Cardiovascular Disease Prevention in Clinical Practice with Representatives of the {{European Society}} of {{Cardiology}} and 12 Medical Societies {{With}} the Special Contribution of the {{European Association}} of {{Preventive Cardiology}} ({{EAPC}})},
  shorttitle = {2021 {{ESC Guidelines}} on Cardiovascular Disease Prevention in Clinical Practice},
  author = {Visseren, Frank L J and Mach, Fran{\c c}ois and Smulders, Yvo M and Carballo, David and Koskinas, Konstantinos C and B{\"a}ck, Maria and Benetos, Athanase and Biffi, Alessandro and Boavida, Jos{\'e}-Manuel and Capodanno, Davide and Cosyns, Bernard and Crawford, Carolyn and Davos, Constantinos H and Desormais, Ileana and Di Angelantonio, Emanuele and Franco, Oscar H and Halvorsen, Sigrun and Hobbs, F D Richard and Hollander, Monika and Jankowska, Ewa A and Michal, Matthias and Sacco, Simona and Sattar, Naveed and Tokgozoglu, Lale and Tonstad, Serena and Tsioufis, Konstantinos P and {van Dis}, Ineke and {van Gelder}, Isabelle C and Wanner, Christoph and Williams, Bryan and {ESC Scientific Document Group}},
  year = {2021},
  month = sep,
  journal = {European Heart Journal},
  volume = {42},
  number = {34},
  pages = {3227--3337},
  issn = {0195-668X},
  doi = {10.1093/eurheartj/ehab484},
  urldate = {2025-03-06}
}

@article{rippeLifestyleStrategiesRisk2019,
  title = {Lifestyle {{Strategies}} for {{Risk Factor Reduction}}, {{Prevention}}, and {{Treatment}} of {{Cardiovascular Disease}}},
  author = {Rippe, James M.},
  year = {2019},
  journal = {American Journal of Lifestyle Medicine},
  volume = {13},
  number = {2},
  pages = {204--212},
  issn = {1559-8284},
  doi = {10.1177/1559827618812395},
  langid = {english},
  pmcid = {PMC6378495},
  pmid = {30800027}
}

@article{sudlowUKBiobankOpen2015,
  title = {{{UK Biobank}}: {{An Open Access Resource}} for {{Identifying}} the {{Causes}} of a {{Wide Range}} of {{Complex Diseases}} of {{Middle}} and {{Old Age}}},
  shorttitle = {{{UK Biobank}}},
  author = {Sudlow, Cathie and Gallacher, John and Allen, Naomi and Beral, Valerie and Burton, Paul and Danesh, John and Downey, Paul and Elliott, Paul and Green, Jane and Landray, Martin and Liu, Bette and Matthews, Paul and Ong, Giok and Pell, Jill and Silman, Alan and Young, Alan and Sprosen, Tim and Peakman, Tim and Collins, Rory},
  year = {2015},
  month = mar,
  journal = {PLoS Medicine},
  volume = {12},
  number = {3},
  pages = {e1001779},
  issn = {1549-1277},
  doi = {10.1371/journal.pmed.1001779},
  urldate = {2025-03-25},
  pmcid = {PMC4380465},
  pmid = {25826379}
}

@misc{UKBiobankUK2025,
  title = {{{UK Biobank}} - {{UK Biobank}}},
  year = {2025},
  month = jan,
  urldate = {2025-03-25},
  howpublished = {https://www.ukbiobank.ac.uk},
  langid = {british}
}

@misc{FraminghamHeartStudy,
  title = {Framingham {{Heart Study}}},
  urldate = {2025-03-25},
  howpublished = {https://www.framinghamheartstudy.org/}
}

@article{thirunavukarasuLargeLanguageModels2023,
  title = {Large Language Models in Medicine},
  author = {Thirunavukarasu, Arun James and Ting, Darren Shu Jeng and Elangovan, Kabilan and Gutierrez, Laura and Tan, Ting Fang and Ting, Daniel Shu Wei},
  year = {2023},
  month = aug,
  journal = {Nature Medicine},
  volume = {29},
  number = {8},
  pages = {1930--1940},
  publisher = {Nature Publishing Group},
  issn = {1546-170X},
  doi = {10.1038/s41591-023-02448-8},
  urldate = {2025-04-14},
  copyright = {2023 Springer Nature America, Inc.},
  langid = {english}
}

@misc{TeachingDatasetsPublic,
  title = {Framingham Teaching {{Datasets}} - {{Public Use Datasets}}},
  urldate = {2025-04-17},
  howpublished = {https://biolincc.nhlbi.nih.gov/teaching/}
}

@misc{ShowcaseHomepage,
  title = {UK Biobank Showcase {{Homepage}}},
  urldate = {2025-04-25},
  howpublished = {https://biobank.ndph.ox.ac.uk/showcase/index.cgi}
}

\clearpage

\section{Methods} \label{supp:method}

Real-world clinical settings pose challenges to CVD risk prediction models, requiring them to handle diverse input information in varying formats and to adapt quickly to different healthcare environments.
To address these requirements, we adopt the paradigm of task-specific fine-tuning of a pre-trained, general-purpose large language model (LLM). This paradigm has demonstrated key properties relevant to our setting:
\begin{enumerate*}[label=(\roman*),itemjoin={\quad}]

    \item transformer-based architectures enable flexible input representations via text prompts and depart from fixed, pre-defined input features;
    \item strong language understanding capabilities allow the incorporation of textual information; and
    \item efficient adaptation capabilities, e.g., through few-shot learning or fine-tuning, support robustness to domain shifts.
\end{enumerate*}

With \textsc{AdaCVD}, we present an approach to employ this paradigm for CVD risk prediction. We use a pre-trained LLM as a starting point and adjust it to the task of CVD risk prediction via supervised, parameter-efficient fine-tuning on real-world data.

%We describe this fine-tuning process in detail in Section \ref{methods:fine-tuning}. 
%We explain the development of this base model in Section \ref{method:base}. To address real-world clinical challenges which we view as domain shifts, we further adapt our base model \textsc{AdaCVD} into three different directions (\textsc{AdaCVD-flex}, \textsc{AdaCVD-text}, and \textsc{AdaCVD-shift}), which we explain in detail in Section \ref{method:adapt}.
%For training and evaluation, we use data from the UK Biobank and the Framingham Heart Study, which we describe in Section \ref{method:data}.

% The following sections detail 
% TODO: mention related work (e.g., MM Generalist Med. AI)

\subsection{Model Architecture and Fine-Tuning} \label{method:training}

Given a patient with individual-specific information, our model predicts the risk of developing CVD within the next 10 years. To achieve this, we fine-tuned LLMs for this specific task in a supervised manner using real-world data. This involves several key components: the choice of base LLM (see Section \ref{methods:LLMs}), the construction of patient prompts (see Section \ref{methods:prompts}), the extraction of risk predictions (see Section \ref{methods:extraction}), and the supervised training process using parameter-efficient fine-tuning (see Section \ref{methods:fine-tuning}).

\subsubsection{Selecting Pre-Trained LLMs as Strong Starting Points} \label{methods:LLMs}
The LLMs we used are all autoregressive, decoder-only transformer models.
We concentrated on open-access LLMs that we can deploy and fine-tune locally to ensure that no sensitive patient data leaves our servers. We focused on two classes of leading open-access LLMs for their balance between performance and computational efficiency during fine-tuning: small models (2-3 billion parameters) and medium-sized models (7-8 billion parameters).
Specifically, we used \texttt{Mistral} (7B) \cite{jiangMistral7B2023}, \texttt{Llama} (3B, 8B) \cite{grattafioriLlamaHerdModels2024}, \texttt{Phi} (3B) \cite{abdinPhi3TechnicalReport2024}, and \texttt{Gemma} (2B) \cite{teamGemmaImprovingOpen2024}. We use the instruction-tuned versions of these models.
Since we observed similar performance after fine-tuning within each model class (see Fig.~\ref{fig:llms-eval}a in Supplementary Material~\ref{supp:results}), we used \texttt{Mistral-7B-Instruct} as the starting point to develop our main model \textsc{AdaCVD}.

\subsubsection{Generating Patient Representations} \label{methods:prompts}

We create two different types of natural language prompts for describing patients: a highly structured one (for \texttt{UKB-structured} and \texttt{Fram}) and an unstructured text that simulates clinical notes (for \texttt{UKB-notes}). Here, we describe how we generated the prompts that we used as input to our model. The data sources used for this process are described in Section \ref{method:data}.

\paragraph{Structured Representations}
To create structured text representations, we serialize data on patients into a detailed textual description, as shown below.

\begin{tcolorbox}[title={\texttt{Structured patient representations}}]
    \texttt{Patient description: \\
    \hspace*{1em} Gender: Male; \\
    \hspace*{1em} Age: 41 years; \\
    \hspace*{1em} <Feature Name>: <Feature Value>; \\
    \hspace*{1em} ... %\\
    %What is the risk that this patient will experience a major cardiovascular event in the next ten years?
    }
\end{tcolorbox}
For all features, we use descriptive and precise names. Depending on the type of feature, the value can be a number (rounded to 1 digit), a short text snippet derived from questionnaire-type information, or a list thereof for questions that allow multiple answers.

\paragraph{Textual Representations}

In the absence of real-world datasets containing unstructured textual descriptions of patients with corresponding 10-year CVD outcomes, we leveraged LLMs to generate patient descriptions that mimic realistic clinical notes. For this, we followed prior work demonstrating LLMs' effectiveness in generating realistic medical summaries \cite{agrawalLargeLanguageModels2022, vanveenAdaptedLargeLanguage2024}.
We used structured patient information as the input and instructed the model to produce a free-text summary of each patient. For the generation, we used two different system prompts, shown below:

\begin{tcolorbox}[title={\texttt{Prompt I for generating patient summaries}}]
    \texttt{You are a medical doctor writing detailed clinical notes. \\ \\
    Patient description: \\
    \hspace*{1em} <Feature Name>: <Feature Value>; \\
    \hspace*{1em} ... \\ \\ 
    Based on this information, generate a concise and natural clinical summary describing the patient in a few sentences.}
\end{tcolorbox}

\begin{tcolorbox}[title={\texttt{Prompt II for generating patient summaries}}]
    \texttt{You are a medical doctor writing detailed clinical notes. \\ \\
    Patient description: \\
    \hspace*{1em} <Feature Name>: <Feature Value>; \\
    \hspace*{1em} ... \\ \\
    Based on this information, generate a brief summary of the patient with an emphasis on relevant cardiovascular-related information. Do not provide risk evaluation or any clinical judgment.}
\end{tcolorbox}

%The generated patient summaries averaged 135 tokens with base risk factors and 248 tokens with additional patient information.
%We capped the lengths at 200 tokens (cropping 3\% of cases) for summaries using only the base risk factors and 400 tokens (cropping 11\% of cases) for the detailed ones. We use the \texttt{Mistral-7B-Instruct} \cite{jiangMistral7B2023} model to generate these summaries. We provide examples of such patient summaries in Supplementary Material \ref{supp:text-representation}.

\subsubsection{Binary Classification in the Token Space} \label{methods:extraction}

To fine-tune LLMs for CVD risk prediction, we framed the problem as a binary classification task in the token space (similar to \cite{hegselmannTabllmFewshotClassification2023, belyaevaMultimodalLLMsHealth2023a}).
Instead of producing a numeric risk prediction in text form, we retrieved the likelihood of the model answering \textit{Yes} or \textit{No} to a question posed in binary form: \emph{Will this patient experience a major cardiovascular event in the next ten years?}
We extracted the logits and subsequently normalized them to generate the final CVD risk prediction.
During training, we completed the prompt with a binary label based on the true observed 10-year CVD outcome of each patient and learned the parameters to minimize the cross-entropy loss between predicted probabilities and observed outcomes. 

\subsubsection{Efficient Fine-Tuning via Low-Rank Adaptation (LoRA)} \label{methods:fine-tuning}

Given the high computational cost of fully training such large models, we employed parameter-efficient fine-tuning (PEFT; \cite{dingParameterefficientFinetuningLargescale2023}), namely Low-Rank Adaptation (LoRA; \cite{huLoRALowRankAdaptation2021}).
LoRA introduces lightweight adapter modules to the attention blocks of the transformer model while keeping the original pre-trained parameters frozen.
Specifically, we targeted the query, key, and value projection layers, with a rank value of 16.
With this approach, we updated only around $0.13\%$ of the model parameters during fine-tuning and thereby significantly reduced computational demands.
The training was done on a cluster of NVIDIA H100 and A100 GPUs.
We provide further details on the training process for all models in Table \ref{tab:training}.

\begin{table}[h]
    \caption{Details on model training and adaptation}
    \label{tab:training}
    \centering
    \begin{tabular}{lllr}
         Model & Initialization & Dataset (size) & \#Epochs   \\
         \toprule
         \textsc{AdaCVD} & \texttt{Mistral-7B-Instruct} & \texttt{UKB-structured} (467k) & 2 \\
         \textsc{AdaCVD-Flex} & \textsc{AdaCVD} & \texttt{UKB-structured} (467k) & 1 \\
         \textsc{AdaCVD-Text} & \textsc{AdaCVD} & \texttt{UKB-notes} (20k) & 5 \\
         \textsc{AdaCVD-Shift} & \textsc{AdaCVD} & \texttt{Fram} (3k) & 10 \\
    \end{tabular}
\end{table}

\subsection{Base Model and Model Adaptation} \label{method:models}

Using the training procedure described in Section \ref{method:training}, we developed a strong base model, AdaCVD, which we further adapted into three directions.
We trained the base model in a setting with complete, structured, and identially distributed data, which is commonly assumed when developing machine learning models. While this setting allows for rigorous evaluation against standard benchmarks, it does not fully reflect the complexities of real-world clinical data. In the clinic, data is often incomplete, unstructured, and subject to distribution shifts across populations (e.g., in different geographic regions). These challenges represent domain shifts. In the following sections, we describe how we developed the base model (Section \ref{method:models-base}) and how we effectively and data-efficiently adapted the base model AdaCVD to handle such shifts (Sections \ref{method:models-adapt-flex} to \ref{method:models-adapt-fram}).

\subsubsection{Training the Core \textsc{AdaCVD} Model} \label{method:models-base}

Following the fine-tuning process outlined in Section \ref{methods:fine-tuning}, we developed our base model \textsc{AdaCVD} using structured representations of patients from the UK Biobank (\texttt{UKB-structured}). We train the model for two epochs on all patients from the training dataset ($n = 467$k), using mini-batches of size 8-16\footnote{The batch size varied depending on the length of the patient descriptions across different settings.}. We provide further technical details on the training process in Table \ref{tab:training}. The hyperparameters were chosen based on the model's performance on the validation set.
%The metrics reported in Section \ref{results} were computed on the test set.

%As established CVD risk scores typically only incorporate a small set of input features (the base risk factors), we first also focus on this setting in order to rigoroulsy evaluate the performance of our model in comparison to the baselines.
%Then, in a second step, we integrate more comprehensive patient information.

%\paragraph{Expert Models}

To assess the importance of different patient information for risk assessment, we trained expert models for each of the 10 information groups defined below (see Section \ref{method:data} for details), each focusing on a different aspect of health-related patient information. For this, we generated patient descriptions solely using the information contained in the specific feature group and the base risk factors. Hence, this process resulted in 11 different expert models: \textsc{Base}, using only the base risk factors; \textsc{Base+X} for the 9 different feature groups; and \textsc{AllPatientInfo}, which uses information from all feature groups simultaneously. Each model was specifically designed to deal with a fixed feature group at inference time.

\subsubsection{Adapting to Incomplete and Variable Patient Information} \label{method:models-adapt-flex}

%\paragraph{A single flexible model}

%To flexibly deal with varying patient information at inference time, ...
Instead of having to deal with 11 different models, depending on what patient information is provided, we aimed to have a single model that could deal with varying information from all feature groups at the same time. For this, we compared two approaches. A straightforward approach is to use \textsc{AllPatientInfo}, i.e., the model trained on complete information of all feature groups, and to simply provide \textit{incomplete} information during inference. Hereby, we assess the models' ability to deal with missing information.
Note, however, that missing values are not explicit \texttt{null} values that require imputation, e.g., with the population median. Instead, incomplete information is only implicit and is simply left out of the patient descriptions.
In a second approach, we made use of the model's adaptability. We adapted the model \textsc{AllPatientInfo} by continuing fine-tuning on patients with varying information. For each patient, we randomly sampled a subset of features during training and generated a patient description using only these features. This way, the model learns to deal more effectively with incomplete information. We refer to the resulting model \ as \textsc{AdaCVD-Flex}.
The performance of the 11 feature expert models provides an upper bound per feature group.

\subsubsection{Adapting to Textual Patient Representations} \label{method:models-adapt-text}

Our initial model, \textsc{AdaCVD}, was fine-tuned at scale on structured patient representations. Even though these representations were encoded in text format, they followed a highly standardized and consistent structure. In contrast, real-world clinical settings rarely provide such uniformity. Patient information is often documented in unstructured formats, such as clinical notes, physician reports, or discharge summaries, making free-text one of the most prevalent data modalities in practice. A key challenge for CVD risk prediction models is thus the ability to process and reason over unstructured text inputs.
Therefore, we conducted an experiment in which we evaluate how well \textsc{AdaCVD}, trained exclusively on structured inputs, generalizes to unstructured text representations in a zero-shot setting. Additionally, we examine how efficiently it can be adapted to this new input format via further fine-tuning, resulting in a variant we refer to as \textsc{AdaCVD-Notes}. For comparison, we also fine-tune the base LLM directly on the textual patient descriptions \textit{from scratch}, without any prior fine-tuning on structured data.

We perform this experiment on our generated dataset of clinical notes (\texttt{UKB-notes}). For some prompts, we provided only the base risk factors as inputs (using Prompt I), and for others, we provided more detailed patient information (all feature groups except lab values, i.e., UA and BS; using Prompt II).
We generate this dataset for a subset ($n=40\,000$) of the UK Biobank cohort. Importantly, we use data from patients not seen during the first fine-tuning stage. To assess the data efficiency, we randomly select subsets for training using different random seeds.

%In particular, we use a subset of the test set for training/adaptation, and report evaluation metrics on the validation set. We do this to ensure that that none of the patients have been seen during the previous fine-tuning stage. 

The generated patient summaries averaged 135 tokens with base risk factors and 248 tokens with additional patient information. We capped the lengths at 200 tokens (cropping 3\% of cases) for base summaries and 400 tokens (cropping 11\% of cases) for detailed ones.

Manual inspection of a subset of the generated summaries confirmed that relevant clinical information was generally preserved, though often rephrased. For example, numerical values were replaced with qualitative descriptors (e.g., \textit{elevated cholesterol levels}), and some features were inferred indirectly (e.g., mentioning obesity instead of stating the BMI value). Summaries based on base risk factors retained nearly all original information, while those including more granular inputs (e.g., physical activity broken down by type and duration) tended to be abstracted (e.g., \textit{the patient is very active}). Our focus in this work does not lie in evaluating these summaries. Instead, we treat them as given and examine how effectively LLMs can learn from such text-based inputs and how efficiently a model trained on structured data adapts to this unstructured format.  Examples of such patient summaries can be found in Supplementary Material \ref{supp:text-representation}.

\subsubsection{Adapting to Distribution Shifts} \label{method:models-adapt-fram}

Distribution shifts are a critical challenge when deploying machine learning models in practice. In clinical decision-making tasks such shifts can occur when applying models across different hospitals, different geographic regions, or temporal contexts. To evaluate how well our model is able to adapt to such shifts, we use data from the Framingham Heart Study. This cohort differs from the UK Biobank in both the time of data collection (Framingham: starting 1948; UKB: starting 2016) and the location (Framingham: US; UKB: UK). We first evaluate how well the base model \textsc{AdaCVD}, trained on UKB data, generalizes to the Framingham dataset without any additional fine-tuning. Then, to assess adaptability, we perform further fine-tuning of \textsc{AdaCVD} using the Framingham data (\textsc{AdaCVD-Shift}). We compare this to fine-tuning the pre-trained LLM \textit{from scratch} only on the Framingham data. To robustly compute metrics, we perform cross-validation using five different random 50-50 train-test splits ($n = 1\,507$ each) and report the median performance. We compare this to the performance of the Framingham risk score, which was derived from a larger cohort of the same study.

\subsection{Data Sources and Cohort Descriptions} \label{method:data} 

We use two primary data sources in this study: the UK Biobank \cite{UKBiobankUK2025,sudlowUKBiobankOpen2015} ($n = 467\,063$) and the Framingham Heart Study \cite{FraminghamHeartStudy} ($n = 3\,014$). 
The UK Biobank (in various formats, as described above) serves as the main dataset for training, adaptation, and evaluation. Data from the Framingham Heart Study is used for adaptation and evaluation.

\subsubsection{UK Biobank}

The UK Biobank is a large-scale longitudinal biomedical database containing detailed health information of over approximately half a million individuals across the UK. It offers a comprehensive repository of patient characteristics, encompassing sociodemographic information, physical measures, lab values, genetic data, lifestyle factors, medical history, and more. Information was collected at a baseline assessment, and after that, disease outcomes and mortality were continuously recorded in a follow-up period of up to 19 years.

\paragraph{Task \& Outcome Definition}

We define our task as predicting the risk of developing a fatal or non-fatal CVD event within 10 years of the baseline assessment.
Hereby, a CVD event is defined as the first occurrence of any of the following ICD-9 and ICD-10 diagnosis codes:
\begin{itemize}
    \item \textbf{ICD-9:} 410--414 (ischemic heart diseases), 430--434 (hemorrhagic and ischemic stroke), and 436--438 (cerebrovascular diseases)
    \item \textbf{ICD-10:} F01 (vascular dementia), I20--I25 (ischemic heart diseases), I50 (heart failure), and I60--I69 (cerebrovascular diseases)
\end{itemize}
This aligns with definitions used in prior studies \cite{alaaCardiovascularDiseaseRisk2019, dagostinoGeneralCardiovascularRisk2008}. 
We combined information from three sources: hospital in-patient admissions, self-reported data, and death registries. Participants with a history of CVD prior to the baseline assessment ($n = 35\,070$) were excluded, applying the same definition for CVD as used for the outcome variable.

\paragraph{Cohort}

The final cohort comprised $467\,063$ participants aged 37--73 years at baseline. The cohort was randomly split into a training ($75\%$), test ($20\%$), and validation set ($5\%$). All reported results are computed on the test set unless stated otherwise. Over the 10-year follow-up period, $7.5\%$ $(n = 34\,983)$ of the participants developed CVD.
Table \ref{tab:cohort} in Supplementary Material \ref{supp:data} shows the baseline characteristics of the study population.

\paragraph{Comprehensive Health Information}

% In contrast to medical risk scores, we ...

We incorporate comprehensive health-related information on individuals and have defined ten distinct information categories designed to reflect realistic clinical scenarios. 
%base risk factors (Base), Lifestyle \& Environment (LE), Sociodemographic factors (SD), Physical Measures (PM), Urine Assays (UA), Blood Samples (BS), Family History (FH), Polygenic Risk Scores (PRS), ICD Codes (ICD), and Medical History (MH).
%10 groups (Base + 9): Explain what information is contained, and what format it has, and how complete it is.
\begin{itemize}
    \item \textbf{Base Risk Factors (Base):} This set of features is commonly used by established CVD risk scores. It consists of age, gender, smoking status, diabetes, total cholesterol, HDL cholesterol, cholesterol medication use, blood pressure, blood pressure medication use, body mass index (BMI), ethnic background, and estimated glomerular filtration rate (eGFR).

    \item \textbf{Polygenic Risk Scores (PRS):} These values quantify the genetic susceptibility of an individual to a broad range of diseases and traits by aggregating the effects of multiple genetic variants. It covers conditions such as cardiovascular diseases, different cancer types, autoimmune disorders, metabolic traits, and neurological and psychiatric disorders. We include 36 scores.

    \item \textbf{Medical History (MH):} Self-reported health information collected through questionnaires, encompassing diagnosed conditions with the individual's age at diagnosis, past medical procedures, medication use, and screening history.

    \item \textbf{Blood Samples (BS):} 43 laboratory-analyzed biomarkers measured in the blood sample collected at recruitment, including 26 biochemistry markers and 17 haematological assays.

    \item \textbf{Family History (FH):} Questionnaire-based information on health conditions of biological and adopted family members, offering insights into hereditary health risks.

    \item \textbf{Lifestyle and Environment (LE):} Self-reported data on physical activity, sleep habits, smoking behavior, and alcohol consumption, providing a comprehensive view of daily routines, health behaviors, and environmental exposures.

    \item \textbf{Physical Measures (PM):} Measurements of body size, body composition by impedance, electrocardiogram (ECG) during exercise, arterial stiffness, and spirometry.

    \item \textbf{Sociodemographics (SD):} Information on living arrangements, household composition, income, education level, employment status, and work conditions.

    \item \textbf{Urine Assays:} Biochemical measurements of urinary components, including creatinine, microalbumin, potassium, and sodium.

    \item \textbf{ICD Codes (ICD):} A record of all past diagnoses using ICD-9 and ICD-10 codes.
\end{itemize}
We provide the exact list of field IDs and features used per category in Table \ref{tab:features} in Supplementary Material \ref{supp:data}.

%\todo{Explain how complete the patient information is per category.}
%The variability across patients arises mainly because not all information is collected for all patients. In some cases, data is conditional on prior responses. For example, detailed questions about past diagnoses were only asked of participants who had reported specific conditions. Thus, shorter descriptions do not necessarily indicate incomplete data entries.

% TODO: Instance-Array. Explain how we process information.

% Importantly, not all features are available for every participant due to the nature of data collection. For example, many questionnaire items are conditional follow-up questions which are only presented to participants with certain conditions. Therefore, complete data across all features is neither expected nor required for every individual.
% To ensure fair comparison of model performance, it is essential to evaluate all models on the same subset of patients. The defined feature groups are sufficiently complete for the vast majority of participants ($\geq95$\%), meaning that most individuals have at least the base features available within each group.

\subsubsection{Framingham Heart Study}

The Framingham Heart Study is a cohort study that focused on cardiovascular disease. It began in 1948 in Framingham, US. The study was instrumental in identifying major CVD risk factors such as hypertension, high cholesterol, smoking, and obesity, and remains one of the most influential epidemiological studies in cardiovascular research. We use the publicly available, de-identified subset of this dataset \cite{TeachingDatasetsPublic}. For each participant, we are given the following information at baseline: Gender, age, total cholesterol, HDL cholesterol, systolic blood pressure, blood pressure medication, diabetes, smoking states, and BMI. We use data from the third period, because previous periods were missing information on HDL cholesterol. We only used participants with complete information. We combine the following three variables as our target: \texttt{CVD}, \texttt{STROKE}, \texttt{MI\_FCHD}. We exclude participants with previous CVD, using the same definition as for the outcome. The final cohort comprised $n=3\,014$ individuals.

\subsection{Evaluation}
\subsubsection{Metrics}

We evaluated the models using standard metrics suitable for unbalanced binary classification tasks. Specifically, we used the area under the receiver operator curve (AUROC) to assess the model's ability to differentiate between individuals who develop the disease and those who do not. Additionally, we used the C-Index to compare the ranking of predictions with the observed time of CVD events. To assess calibration, we used the Brier Score.
%For all metrics, we reported the median value and its $95\%$ confidence interval, computed using bootstrapping with $5000$ bootstrapping rounds. % Uncertainty about the estimate (median; ci)
For all metrics, we report the median value and their $95\%$ spread across $5000$ bootstrapping rounds. % Spread of the data (pi)
The observed large spreads are a result of high sample dependence, which is likely due to class imbalance. We decided not to measure randomness across different training runs (e.g., different seeds or initialization parameters) due to the high computational cost. However, we observed very stable results with respect to such randomness.

\subsubsection{Comparisons \& Baselines}

For all experiments using tabular input features, we compared our method with various baseline methods, including medical risk scores, standard machine learning methods, and LLMs (zero-shot). 

\paragraph{Medical Risk Scores}
We implemented medical risk scores derived from different geographic cohorts. We list all risk scores, geographic regions, and the exact sets of input features in Table~\ref{tab:risk_scores}.

% age, gender, smoking status, diabetes, total cholesterol, HDL cholesterol, cholesterol medication use, blood pressure, blood pressure medication use, body mass index (BMI),  ethnic background, and estimated glomerular filtration rate (eGFR)
\begin{table}[h] 
\caption{Medical Risk Scores}
\label{tab:risk_scores}
    \centering
    \begin{tabular}{llp{3in}}
    \toprule
    \textbf{Risk Score} & \textbf{Derivation Cohort} & \textbf{Input Features} \\
    \midrule
    Framingham \cite{dagostinoGeneralCardiovascularRisk2008} & US & Age, sex, total cholesterol, HDL cholesterol, systolic BP, BP medication, smoking status, diabetes  \\
    PREVENT \cite{khanDevelopmentValidationAmerican2024} & US & Age, sex, total cholesterol, HDL cholesterol, systolic BP, BP medication, smoking status, diabetes, eGFR, cholesterol medication, BMI \\
    ASCVD (AHA/ACC) \cite{arnett2019ACCAHA2019} & US & Age, sex, total cholesterol, HDL cholesterol, systolic BP, BP medication, smoking status, diabetes, ethnicity  \\
    SCORE2 \cite{score2workinggroupandesccardiovascularriskcollaborationSCORE2RiskPrediction2021} & Europe & Age, sex, total cholesterol, HDL cholesterol, systolic BP, smoking status, diabetes \\
    QRISK \cite{hippisley-coxDerivationValidationQRISK2007} & UK & Age, sex, total cholesterol, HDL cholesterol, systolic BP, BP medication, smoking status, diabetes, townsend deprivation index\footnote{Not provided in our dataset; we impute it with the median value of ...}, family history of premature CVD, BMI
    \end{tabular}
\end{table} 

\paragraph{Machine Learning Baselines}
The second group of baseline models comprises standard supervised machine learning methods, including the Cox Proportional Hazards model, logistic regression, and gradient-boosted trees. %For each model, we chose optimal hyperparameters by searching over a wide search space (see Table~\ref{tab:search_space}).
We used the following software packages for the implementations: \texttt{lifelines} for the Cox PH model, \texttt{sklearn} for logistic regression, and \texttt{lightgbm} for gradient-boosted trees.

\paragraph{LLMs (Zero-Shot)}

To assess the zero-shot predictions of different pre-trained LLMs, we provided a patient description, gave precise instructions, and extracted the prediction from the response, similar to \cite{hanEvaluationGPT410year2024a}. We instructed the models to utilize a JSON format within their responses to ensure straightforward extraction of the numeric risk prediction. Specifically, our instruction was: \textit{Based on the provided patient description, what is the estimated 10-year risk of cardiovascular disease (CVD)? Please provide your answer solely as a numeric percentage in a machine-readable JSON format.}
We generated 100 new tokens and extracted the risk prediction from the response. If no valid JSON was provided, we set the prediction to \texttt{nan}. When a model did not comply with the instructions, all predictions were invalid and hence, we were not able to compute any evaluation metrics. 
\section*{Code Availability}
The source code for \textsc{AdaCVD} is available
at \url{https://github.com/FrederikeLuebeck/adacvd}.

\section*{Acknowledgments}

This work was carried out under UK Biobank application number 60520.

\section*{Author contributions}

F.T. and S.G. initiated and conceived the study;
F.L., J.W., F.T., S.G., A.K., and B.S. contributed to the refinement of the research question;
F.L., J.W., and F.T. devised the model architecture and evaluation framework;
F.L. and S.G. performed the dataset curation and data analysis; F.L. performed data processing, model training, evaluation, and all experiments;
M.M. reviewed the code;
F.L., J.W., F.T., S.G., A.K., and B.S. wrote the manuscript; 
All authors approved the final version of the manuscript.

\section*{Declaration of interests}

The authors declare no competing financial interests.

%\end{multicols}

\clearpage
\newpage
\appendix

\section{Supplementary Material}

\subsection{UK Biobank} \label{supp:data}

\begin{table*}[h] 
\caption{Characteristics of the UK Biobank Cohort, excluding participants with CVD prior to the baseline assessment. We report median values and their standard deviation.}
\label{tab:cohort}
    \centering
    \begin{tabular}{lrr}
        \toprule
        % 207816.0
        \textbf{} &\textbf{Female (n = 261\,030)} & \textbf{Male (n = 206\,033)}  \\
        \midrule
        \textbf{Age (years)} & 57.00 (7.99) & 57.00 (8.21) \\ 
        \textbf{BMI (kg/m²)} & 26.03 (5.14) & 27.18 (4.17) \\ 
        \textbf{Total Cholesterol} & 225.99 (43.09) & 214.66 (42.13) \\ 
        \textbf{HDL Cholesterol} & 60.33 (14.58) & 48.26 (12.03) \\ 
        \textbf{Systolic Blood Pressure} & 133.00 (19.23) & 139.50 (17.38) \\ 
        \textbf{Blood Pressure Medication} & 15.80\% & 19.76\% \\
        \textbf{eGFR} & 97.60 (13.01) & 97.53 (12.69) \\ 
        \textbf{Smoker} & 8.78\% & 12.39\% \\ 
        \textbf{Diabetic} & 3.37\% & 5.74\% \\
    \end{tabular}
\end{table*} 

\begin{table*}[h]
  \centering
  \caption{List of field IDs used for the information categories in the UK Biobank. IDs marked with an asterisk are further processed into features. Information on the field can be found on the UK Biobank Showcase Webpage~\cite{ShowcaseHomepage}.}
  \label{tab:features}
    \begin{tabularx}{1.0\textwidth}{l>{\raggedright\arraybackslash}X}
        \toprule
        ~ & \textbf{Field IDs} \\ \midrule \midrule
        \textbf{Base} & 31, 93, 2443, 4080, 6153$^*$, 6177$^*$, 20116, 21000, 21001, 21003, 30690, 30700$^*$, 30760 \\ \midrule
        \textbf{PRS} & 26202, 26204, 26206, 26210, 26212, 26214, 26216, 26218, 26220, 26223, 26225, 26227, 26229, 26232, 26234, 26238, 26240, 26242, 26244, 26246, 26248, 26250, 26252, 26254, 26258, 26260, 26265, 26267, 26269, 26273, 26275, 26278, 26283, 26285, 26287, 26289 \\ \midrule
        \textbf{MH} & 2178, 2188, 2296, 2306, 2316, 2345, 2355, 2365, 2415, 2443, 2453, 2463, 2473, 2492, 2844, 2966, 2976, 3005, 3761, 3786, 3809, 3992, 4012, 4022, 4041, 4717, 6150, 6151, 6152, 6153, 6154, 6155, 6177, 6179 \\ \midrule
        \textbf{BS} & 23000, 30000, 30010, 30020, 30030, 30040, 30050, 30060, 30070, 30080, 30090, 30100, 30110, 30120, 30130, 30140, 30150, 30160, 30600, 30610, 30620, 30630, 30640, 30650, 30660, 30670, 30680, 30690, 30700, 30710, 30720, 30730, 30740, 30750, 30760, 30770, 30780, 30790, 30810, 30840, 30860, 30870, 30880, 30890 \\ \midrule
        \textbf{ICD} & 41280$^*$, 41270$^*$, 41281$^*$, 41271$^*$ \\ \midrule
        \textbf{FH} & 1807, 1845, 3526, 4501, 20107, 20110, 20111, 20112, 20113, 20114 \\ \midrule
        \textbf{SD} & 670, 709, 728, 738, 767, 777, 796, 806, 816, 826, 845, 3426, 4674, 6138$^*$, 6143, 20119 \\ \midrule
        \textbf{LE} & 864, 874, 884, 894, 904, 914, 924, 943, 971, 981, 991, 1001, 1011, 1021, 1070, 1080, 1090, 1160, 1190, 1200, 1210, 1220, 1239, 1249, 1259, 1269, 1279, 1558, 1568, 1578, 1588, 1598, 1608, 1618, 1628, 2624, 2634, 3637, 3647, 20116, 20117, 20160, 20161, 20162, 22035, 22036, 22037, 22038, 22039 \\ \midrule
        \textbf{PM} & 3062, 3063, 3064, 4194, 4195, 4196, 4198, 4199, 4200, 4204, 4207, 5983, 6015, 6016, 6017, 6032, 6033, 6034, 6039, 20150, 20151, 20256, 20257, 20258, 21001, 21021, 23098, 23099, 23100, 23101, 23102 \\ \midrule
        \textbf{UA} & 30500, 30505, 30510, 30520, 30525, 30530, 30535 \\ \bottomrule
    \end{tabularx}
\end{table*}

\clearpage
\subsection{Examples of textual patient representations} \label{supp:text-representation}

%Below, we provide randomly selected examples.

% Examples
\begin{tcolorbox}[]
    The patient is a 41-year-old non-smoking, non-diabetic female of white ethnicity with a BMI of 23.1 Kg/m2. She has a cholesterol level of 4.9 mmol/L and an HDL cholesterol level of 1.9 mmol/L. Her blood pressure, as measured automatically, is 108.5 mmHg. She is not currently taking any cholesterol-lowering medication or blood pressure medication. Her eGFR is 120.37, indicating normal kidney function.
\end{tcolorbox}

\begin{tcolorbox}[]
    The patient is a 61-year-old male with a BMI of 24.9 Kg/m2, previously a smoker but not currently. He has a cholesterol level of 4.8 mmol/L and a low HDL cholesterol level of 1.1 mmol/L. His systolic blood pressure is 133 mmHg. He does not have diabetes, is not on blood pressure medication, and does not take cholesterol-lowering medication. His estimated glomerular filtration rate (eGFR) is 77.55, indicating good kidney function.
\end{tcolorbox}

\begin{tcolorbox}[]
    The patient is a 48-year-old non-smoking, non-diabetic female of white ethnicity with a normal body mass index (BMI) of 21.9 Kg/m2. She has a borderline high cholesterol level, with a low HDL cholesterol level. Her blood pressure, as measured automatically, is slightly elevated at 134.5 mmHg. She is not currently on any cholesterol-lowering medication or blood pressure medication. Her estimated glomerular filtration rate (eGFR) is within the normal range at 96.05.
\end{tcolorbox}

\begin{tcolorbox}[]
The patient is a 41-year-old female with a BMI of 23.1 Kg/m2, who has never smoked and has no history of diabetes or hypertension. Her cholesterol level is 4.9 mmol/L, with an HDL cholesterol of 1.9 mmol/L. She is currently not on any cholesterol-lowering medication. Her systolic blood pressure, as measured automatically, is 108.5 mmHg. She has a family history of non-accidental death in close genetic family members. Her PRS for cardiovascular disease (CVD) is 2.3 relative risk, and her PRS for venous thromboembolic disease (VTE) is 2.2 relative risk. She engages in regular walking and light DIY, and her sleep duration is 8 hours/day. She consumes alcohol three or four times a week, with an average weekly spirits intake of 4 measures. Her maximum workload during a fitness test was 80 Watts, and her maximum heart rate during the test was 139 bpm. She lives in a house or bungalow with 2 people and has a college or university degree as her highest qualification.
\end{tcolorbox}

\begin{tcolorbox}[]
The patient is a 65-year-old female with a BMI of 22.9 Kg/m2. She is a current smoker and has a systolic blood pressure of 122 mmHg. Her cholesterol level is 6.1 mmol/L, with an HDL cholesterol of 1.3 mmol/L. She is not taking any cholesterol-lowering medication. Her estimated glomerular filtration rate (eGFR) is 89.92. She has a standard polygenic risk score (PRS) for coronary artery disease (CAD) of 1.1 relative risk. She has no history of diabetes, hypertension, or cardiovascular disease. She is physically active, walking 7 days a week and engaging in moderate physical activity for 300 minutes a day. She has no known vascular or heart problems diagnosed by a doctor. Her sleep duration is 6 hours a day, and she does not snore or daytime doze. She has a standard PRS for hypertension of 0.3 relative risk.
\end{tcolorbox}

\clearpage
\subsection{Additional Results} \label{supp:results}

\begin{figure*}[h!]
    \centering
    \includegraphics[width=\textwidth]{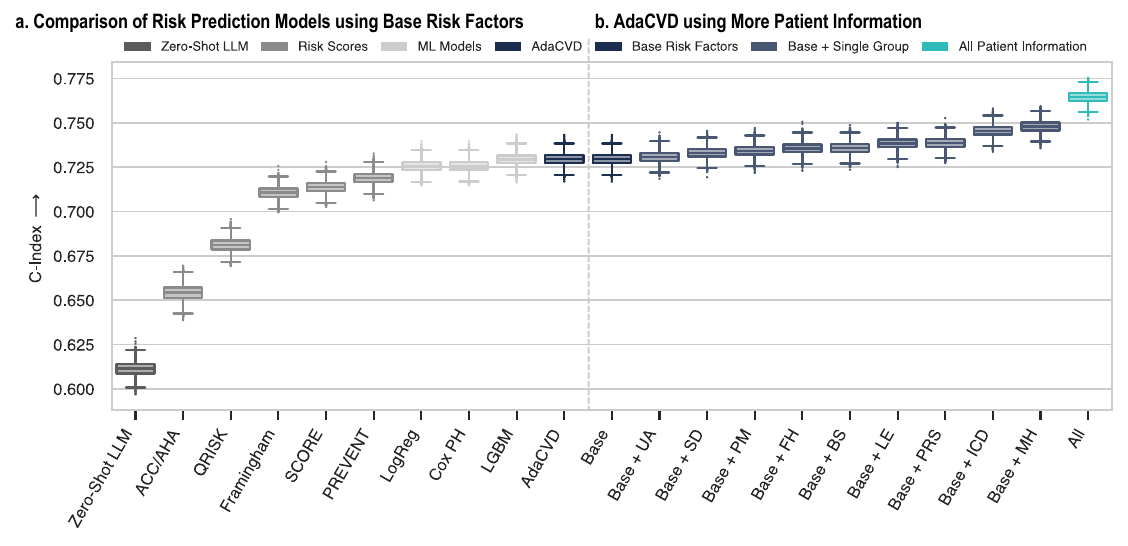}
    \caption{\textbf{Evaluation of predictive performance of different risk prediction models (C-Index).}
    \textbf{a.}~Comparison of CVD risk prediction models using only a limited set of base risk factors. 
    \textbf{b.}~\textsc{AdaCVD} when additional patient information is incorporated.
    Acronyms: Urine Assays (UA), Sociodemographic factors (SD), Physical Measures (PM), Family History (FH), Blood Samples (BS), Lifestyle \& Environment (LE), Polygenic Risk Scores (PRS), ICD Codes (ICD), and Medical History (MH). Each feature group includes the base risk factors. The \textit{All Patient Information} setting integrates all feature categories.
    }
    \label{fig:supp-cindex}
\end{figure*}

\begin{figure*}[h!]
    \centering
    \includegraphics[width=\textwidth]{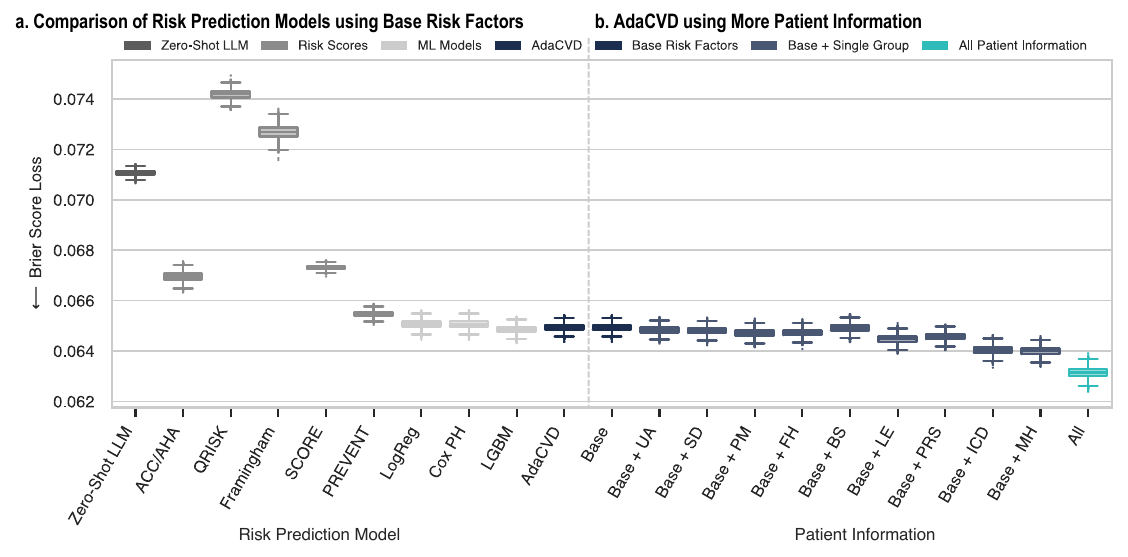}
    \caption{\textbf{Evaluation of calibration of different risk prediction models (Brier Score Loss).}
    \textbf{a.}~Comparison of CVD risk prediction models using only a limited set of base risk factors. 
    \textbf{b.}~\textsc{AdaCVD} when additional patient information is incorporated.
    Acronyms: Urine Assays (UA), Sociodemographic factors (SD), Physical Measures (PM), Family History (FH), Blood Samples (BS), Lifestyle \& Environment (LE), Polygenic Risk Scores (PRS), ICD Codes (ICD), and Medical History (MH). Each feature group includes the base risk factors. The \textit{All Patient Information} setting integrates all feature categories.
    }
    \label{fig:supp-brier}
\end{figure*}

\begin{figure*}[h]
    \centering
    \includegraphics[width=\textwidth]{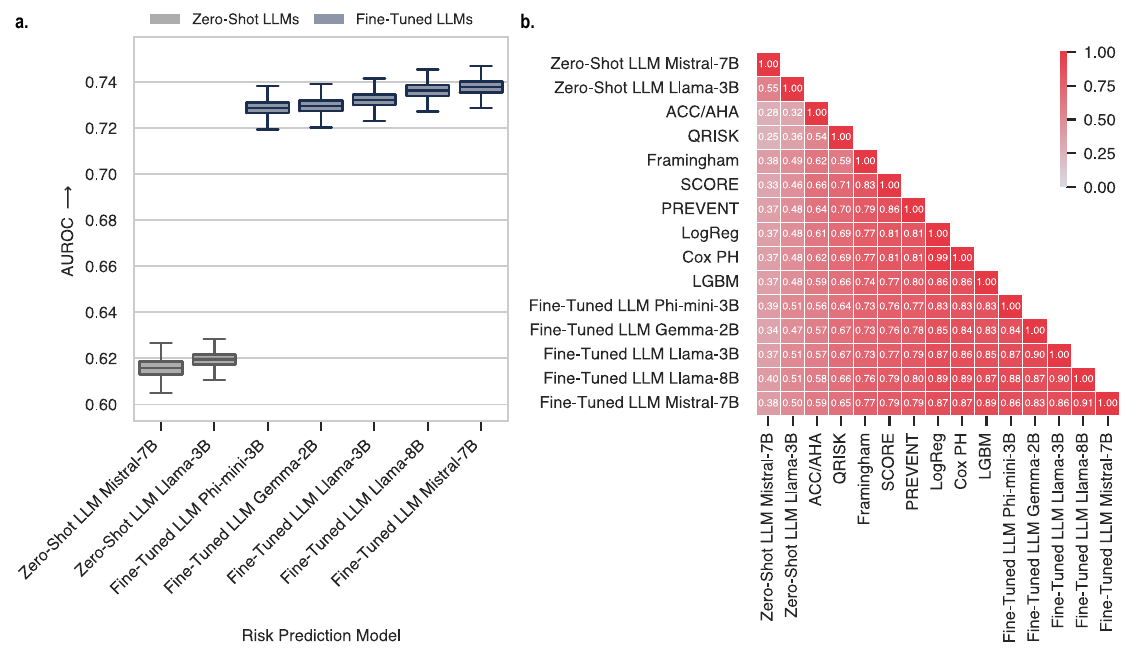}
    \caption{\textbf{Evaluation of risk prediction models using the base risk factors.}
    \textbf{a.}~Comparison of different LLMs of small and medium size, both zero-shot and fine-tuned. LLMs not shown in the zero-shot group did not comply with the instructions.
    \textbf{b.}~Correlation between predictions of different risk prediction models, as measured by the Kendall rank correlation coefficient.
    }
    \label{fig:llms-eval}
\end{figure*}

\clearpage

\begin{figure*}[h!]
    \centering
    \includegraphics[width=\textwidth]{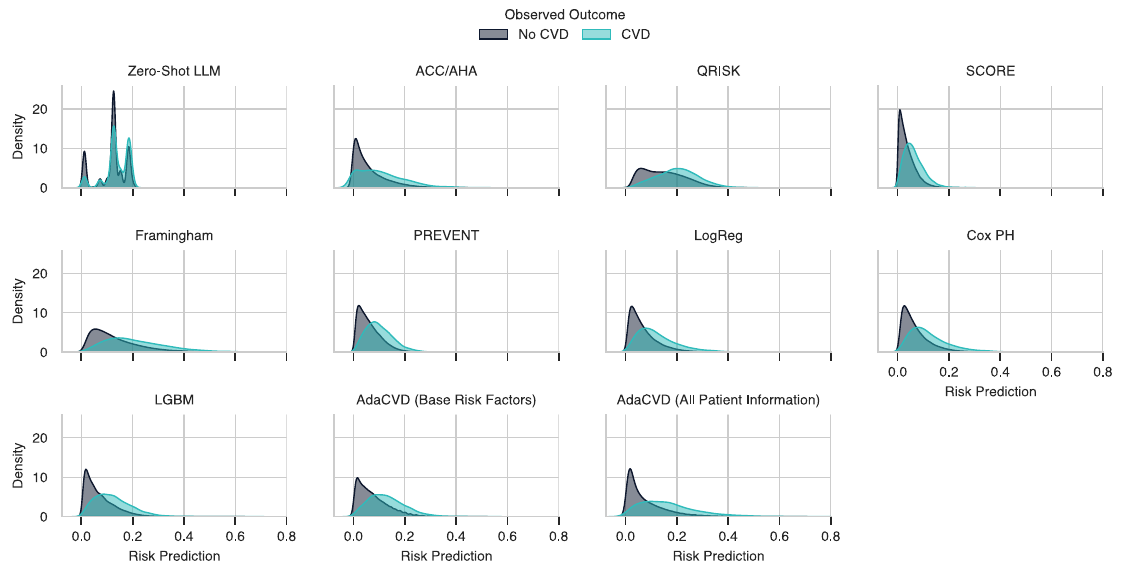}
    \caption{\textbf{Distribution of risk predictions.} Risk prediction distributions for individuals that developed CVD vs. those who did not for different risk prediction models.
    }
    \label{fig:supp-distribution}
\end{figure*}

\begin{figure*}[h!]
    \centering
    \includegraphics[width=\textwidth]{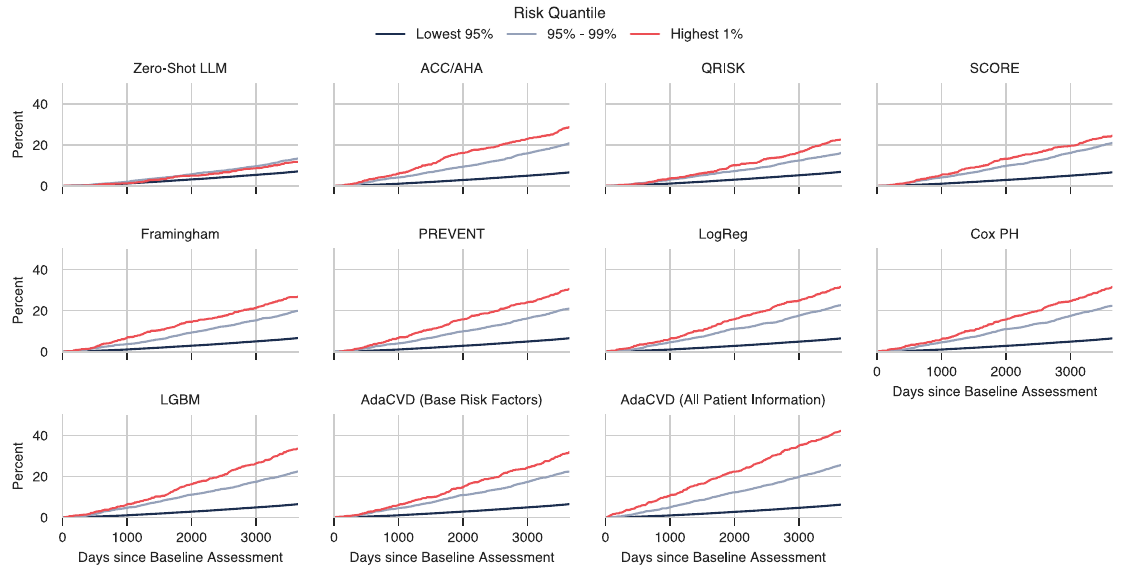}
    \caption{\textbf{Stratified event curves. }~Event curves stratified by predicted risk percentiles (lowest $95\%$, $95-99\%$, and highest $1\%$) for different risk prediction models. The x-axis denotes the 10-year follow-up period (in days); The y-axis shows the observed incidence for each risk group up until that day; The rightmost points indicate the observed 10-year incidence rates for each risk group.
    }
    \label{fig:supp-curves}
\end{figure*}

\begin{comment}
\begin{figure*}[h]
    \centering
    \includegraphics[width=0.6\textwidth]{figures/fig-llm-lgbm.pdf}
    \caption{...}
    \label{fig:llm-lgbm}
\end{figure*}
\end{comment}

\begin{figure*}[h]
    \centering
    \includegraphics[width=\textwidth]{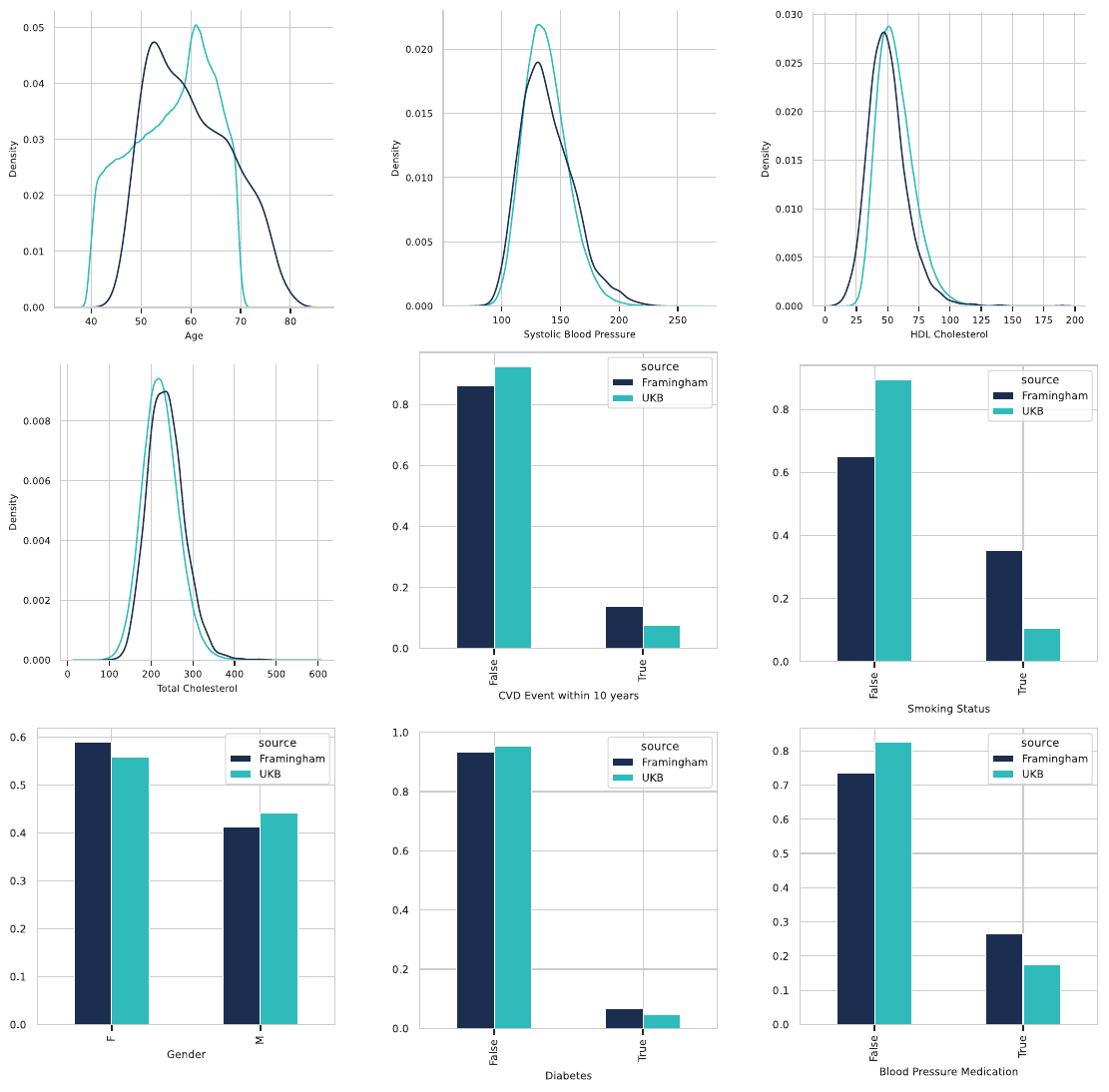}
    \caption{Comparison of the marginal distributions of the base risk factors between the UK Biobank and the Framingham Heart Study participants.}
    \label{fig:ukb-fram}
\end{figure*}

\end{document}